\documentclass{article}
\usepackage{iclr2025_conference,times}

\usepackage{amsmath,amsfonts,bm}

\def\eqref#1{equation~\ref{#1}}

\def\1{\bm{1}}

\DeclareMathAlphabet{\mathsfit}{\encodingdefault}{\sfdefault}{m}{sl}
\SetMathAlphabet{\mathsfit}{bold}{\encodingdefault}{\sfdefault}{bx}{n}

\usepackage{amsmath,amssymb, comment}
\usepackage{graphicx, hyperref}
\usepackage{float}
\usepackage{xcolor}
\usepackage{url}
\iclrfinalcopy          

\title{AllocBench: Measuring Online Tool Allocation Capability in LLM Agents}

\author{%
\makebox[\textwidth]{%
\begin{tabular}[t]{c}
Daniel Wang\thanks{These authors contributed equally to this paper.} \\
Yale University \\
\texttt{daniel.wang.dzw22@yale.edu}
\end{tabular}
\hspace{0.08\textwidth}
\begin{tabular}[t]{c}
Andrew Xu\footnotemark[1] \\
Yale University \\
\texttt{andrew.xu.ax59@yale.edu}
\end{tabular}%
}}

\begin{document}
\maketitle
\begin{abstract}
Creating a reusable tool is an investment: an agent pays a fixed cost now in
exchange for the potential of future reuse. Therefore, a user should prefer an agent that creates a small number of highly reusable tools, rather than many one-offs. We introduce a paired benchmark that tests whether LLM agents exhibit conscious allocation behavior under a fixed budget in two contexts: an abstract text-based formulation and a code-construction task. We find that every frontier model we test---Claude Haiku, Claude Opus, GPT-5.4-mini, and GPT-5.6 Sol---acts near-optimally in the abstract framing but fails to transfer this ability to script-writing. Through further experiments, we
identify the particular failure modes for each model. Notably, the first three models fail even when the scripts are not evaluated, while GPT-5.6 Sol stays selective under that weaker manipulation and collapses only at full construction. Furthermore, an open-source Qwen model policy-trained for abstract allocation generalizes this ability across held-out lexical
variations, but sees no improvement at script allocation.
Together, these results establish online tool allocation as a significant
capability boundary, even for modern frontier models.
\end{abstract}

\section{Introduction}

Large language model (LLM) agents are increasingly used to create scripts, wrappers, macros, and skills that can be reused across tasks. Existing evaluations ask whether a model can
write a correct tool, choose an available tool, or select from a growing
library \citep{cai2024latm,qian2023creator,huang2024ultratool,qu2025toollearningsurvey}. 
They largely leave a more fundamental decision unmeasured: \emph{when} should the agent create a tool? 

Building tools---API calls, Python scripts, or video game macros---is expensive, consuming both time and tokens. For individual tasks, paying this cost is almost always either obviously worthwhile or clearly unnecessary. On the other hand, agents that may expect to see certain tasks recur must take this tradeoff more seriously: tool creation incurs a fixed cost but makes future uses far cheaper. An agent facing an unknown task stream therefore has two
options at each turn: write a script for the current task or wait for more evidence of recurrence. Building too eagerly wastes resources on rare tasks that could have been better treated by hand; waiting too long forfeits useful reuse. 

The cost of tool construction is difficult to pinpoint. For example, the number of tokens needed for a model to generate a Python script depends on the capability of the model and the complexity of the task. This leaves providers of agentic services in a difficult position---in principle, the model itself should decide when building is beneficial, but recent work has shown that current models fail to do so reliably \citep{lin2026bagen}. An alternative approach to encourage more prudent tool usage is to present the agent with a limited action budget \citep{liu2025budgetaware}. One might hope and expect that decreasing the number of scripts an agent is permitted to write over a stream of tasks would make it more selective with its tool creation. We show in this paper that this fails to hold, even in the most capable modern frontier models.

A small but growing set of studies---many from the last year---have analyzed how LLM agents respond to budget awareness. A consensus seems to be that models are naturally poor at budgeting, but that this capability can be easily post-trained through a variety of techniques \citep{liu2025budgetaware, lin2026bagen, li2026bavt, ding2026calibrate}. We find this assessment only partially accurate. By introducing a paired benchmark to test budget-aware tool investment, we find that a significant capability wall exists at the gap between abstract allocation and tool creation, which has not yet been addressed in a budget-aware context. Furthermore, policies learned in the former do not always transfer to the latter, revealing a fundamental and unexpected failure mode in modern LLMs. 

In this paper, we use an experimental structure that intentionally generalizes to a variety of domains. In one session, we present the agent with a sequence of $T$ tasks chosen from a hidden distribution. The agent can create at most $B$ tools throughout the entire session, and tools can be reused once made. Because tasks and tools can take any form, this framework can be applied to a wide range of settings; we consider an abstract ball-collection game and a concrete math test. This is an online learning problem, where optimal behavior involves the agent basing its actions based on the already-observed distribution.

Our primary contributions are as follows:
\begin{enumerate}
    \item a generalizable paired benchmark to study budget-aware tool investment, which we call AllocBench;
    \item an accompanying numerical score to fairly compare performance across models;
    \item strong cross-family evidence that frontier models can abstractly allocate budget but fail to do so in tool creation;
    \item a localization of the trigger, which sits at required code emission for three of the four frontier models and at full script construction for the strongest;
    \item a demonstration that post-training on abstract budget allocation can fail to transfer to tool creation;
    \item a conjecture that this capability suppression is caused by certain characteristics of predominant post-training methods.
\end{enumerate}

\section{Related Work}

\paragraph{Tool use benchmarks.} LLM tool creation has been evaluated in many varied contexts. The first dataset introduced in this direction was the Creation Challenge \citep{qian2023creator}, which prompts models to write Python scripts to solve algebra word problems. Later benchmarks have primarily focused on simulating realistic settings \citep{huang2024ultratool, yu2026wildtoolbench}. These benchmarks test tool creation, use, and reuse over lengthy problem streams. Numerous methodologies have also been introduced to improve LLM tool capabilities. LATM uses a larger model to write scripts and a smaller model to reuse them
\citep{cai2024latm}; other works such as ReGAL, LILO, and TroVE all induce expandable skill libraries and show improvements over one-shot tool creation \citep{stengeleskin2024regal,grand2024lilo,wang2024trove}. These systems assume that the recurring task warrants construction. In fact, recent work shows that these methods result in much less reuse than expected, instead improving self-correction
\citep{berlotattwell2024librarylearning}. Therefore, an important and necessary question to answer before focusing on inducing reuse is whether LLMs can find the right opportunities to take advantage of it.

\paragraph{Budget awareness.} A separate and very recent line of work asks whether agents respond to a fixed budget (primarily of tool calls) at inference time. Budget-Aware Tool-Use finds that granting a larger tool-call budget does not improve web-search agents because they lack native budget awareness \citep{liu2025budgetaware}. The same paper introduces a prompt-level budget tracker and a
budget-aware framework that improves cost efficiency. Similarly, BAGEN evaluates whether agents can estimate their own remaining budget mid-episode, finding that strong task
performance and capable budget-awareness are only weakly correlated, and even frontier models are over-optimistic \citep{lin2026bagen}. Several other methodologies of increasing budget awareness involve computing the expected utility gain from calling a tool \citep{liu2026utility}, pruning a reasoning tree \citep{li2026bavt}, and planning to maximize the probability of finishing within a deadline and under budget \citep{wang2026mcpp}. This last work studies ``online resource allocation'' by allowing models to allocate compute across
subtasks of a single workflow. Though the terminology is very similar to this paper, it studies a very different problem from tool investment. Notably, none of these studies provide a quantitative metric for budget allocation, and none provide cross-model comparisons. By discretizing the action space with a defined problem stream, we fill both these gaps.

\paragraph{Online learning and regret.} From a theoretical standpoint, AllocBench best aligns with prior work on online learning capability in LLMs. We draw inspiration from the theory of LLM \textit{regret} \citep{park2025regret}, which measures the expected reward the agent failed to obtain. LLMs are known to have non-zero regret even on simple tasks \citep{park2025regret}. Regret has been used as an unsupervised training loss, a method that is proven in theory to improve LLM performance \citep{park2025regret}. In this paper, we employ a variation of regret as a post-training loss in reinforcement learning (RL). Our results agree with studies that have demonstrated suboptimal behavior stemming from premature commitment to a strategy \citep{chen2025greedy, mehta2026commit}. The recently-introduced Calibrate-Then-Act framework seeks to prevent this by explicitly providing the agent with an estimated prior about its environment state \citep{ding2026calibrate}.  Separately, small variations in problem structure (e.g. arithmetic in base 9 vs. base 10) have been shown to significantly degrade LLM performance \citep{wu2024counterfactual}. We present the first benchmark designed to test online budget allocation, as well as the first framing gap for online learning.

\section{AllocBench}
\label{sec:allocbench}

\subsection{Formulation}
\label{sec:decision-process}

An episode or ``stream'' of AllocBench consists of a hidden sequence of $T$ instances from $N$ latent classes, which we denote $c_1,\ldots,c_T\in \{1,\ldots,N\}.$ The stream generator is described in Section~\ref{sec:construction}. Let $M_t$ (with $M_1=\emptyset$) denote the set of classes the agent holds an existing tool for at turn $t$, and let $b_t=B-|M_t|$ be the remaining tool budget. At each turn the agent observes $o_t = (x_t, b_t, t, h_t)$: a frame-dependent rendering $x_t = \rho(c_t, \theta_t)$ of the current class with fresh parameters $\theta_t$; the remaining budget; the turn index; and the full interaction history, denoted $h_t$. Class identity is derivable from $x_t$ as described in Section~\ref{sec:frames}. If $c_t \notin M_t$ and $b_t > 0$, the agent chooses $a_t \in \{\textsc{commit}, \textsc{pass}\}$. Picking \textsc{commit} sets $M_{t+1} = M_t \cup \{c_t\}$, $b_{t+1} = b_t - 1$, and credits the current occurrence; \textsc{pass} forfeits it. If $c_t \in M_t$, the artifact is applied automatically and the occurrence credited with no decision. If $b_t = 0$, only \textsc{pass} remains.\footnote{In Appendix~\ref{app:economic}, we describe a variant of this formulation in which committing incurs an additional disclosed cost.} Each credited occurrence earns one point. For a policy $\pi$ mapping observations $o_t$ to actions $a_t$, the per-turn reward $r_t$ and the total utility $U(\pi)$ are defined by 
\begin{equation}
r_t = \mathbb{I}\{c_t \in M_{t+1}\},
\qquad
U(\pi) = \sum_{t=1}^{T} r_t.
\label{eq:utility}
\end{equation}

Intuitively, committing budget on the first instance of a class is rarely optimal, as waiting for more information loses only one occurrence while budget spent on a rare class forfeits an entire frequent class in opportunity cost. Appendix \ref{app:policy-math} shows that under a symmetric Dirichlet prior, an agent benefits significantly on expectation by waiting for a class to repeat before committing.  We design the test streams to amplify this effect by guaranteeing an early trap (see Section \ref{sec:construction}). Figure~\ref{fig:optimal-policy} depicts the optimal online behavior on a real AllocBench episode; the optimum passes every first instance and builds on confirmed recurrence.

\subsection{Stream generation}
\label{sec:construction}

\begin{figure}
\centering
\includegraphics[width=\linewidth]{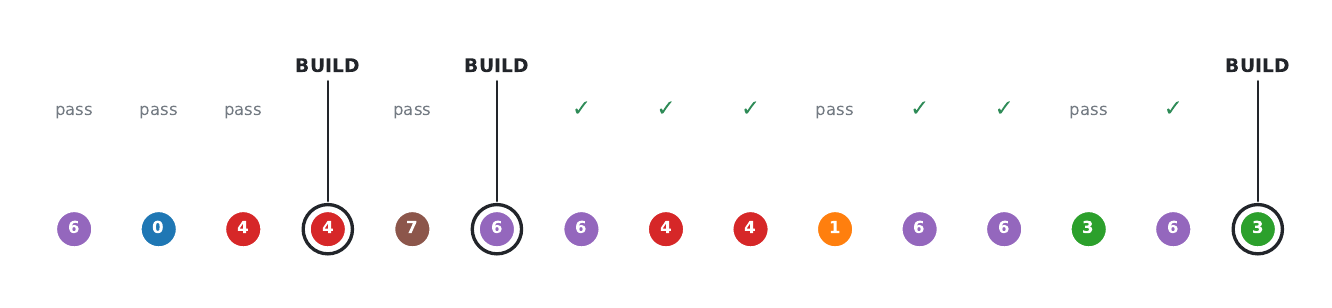}
\caption{\textbf{The optimal policy: pass on first sight, build on confirmed recurrence.} Colored circles represent the first 15 of 60 classes from one canonical stream (seed 2000; see Appendix \ref{app:reconstruction}). The policy spends the $B=3$ budget on the three hot classes and collects later occurrences of built classes for free; the trap classes are correctly left unclaimed.}
\label{fig:optimal-policy}
\end{figure}

AllocBench uses streams with $T = 60$ numeric word problems drawn from $N= 8$ families: linear congruential generators, modular exponentiation, continued fractions, Chinese-remainder systems, Josephus survivor, iterated modular quadratic maps, xorshift steps, and modular matrix powers. Each occurrence is rendered with fresh parameters and a randomized cover story at a configurable magnitude (see Appendix \ref{app:reconstruction} for details). For each episode, three families are designated \emph{hot} and five \emph{trap}, uniformly at random. Hot families share $0.85$ of the probability mass ($0.28\overline{3}$ each) and traps $0.15$ ($0.03$ each), so hots recur roughly 13--23 times and traps 1--6 times. To make it easier to distinguish between online allocation policies (e.g. Figure \ref{fig:optimal-policy}) and an eager policy that exhausts budget immediately, the sequence is rejection-sampled until at least one trap occurs within the first $B=3$ slots. We verify that all tested models are approximately unable to solve the problems without a script (see Section~\ref{sec:gate}), making budget allocation genuinely important. We generate 24 independent streams (labeled seeds 2000--2023) as the evaluation panel.

\subsection{Framing ladder}
\label{sec:frames}

Each stream is rendered at one of four framing rungs, ordered from abstract to fully constructive:
\begin{description}
\item[R0 (abstract):] the model sees differently colored balls and chooses to \textsc{keep}/\textsc{pass} in free-text.
\item[R1 (declarative):] real numeric problems with declarative \textsc{keep}/\textsc{pass} tool calls, no code.
\item[R2 (code):] keeping an instance requires a non-empty \texttt{solve(inputs)} implementation; payoff is independent of code correctness, though the model is still instructed to write correct code.
\item[R3 (full):] reusable-script construction, tested and rerun on later instances; the model may instead attempt a raw answer, credited only if correct.
\end{description}
All rungs render identical latent sequences, so the defer-then-confirm policy in Figure~\ref{fig:optimal-policy} remains optimal at every rung. We note that while class identity is immediate in R0 and must be inferred in later rungs, our experiments show that cross-type misapplication is very rare and unlikely to cause meaningful variations in performance  (Appendix~\ref{app:mech-validity}).

\subsection{Metrics and scoring}
\label{sec:outcomes}

In Section~\ref{sec:decision-process} we argued that committing on a class's first instance is rarely optimal. To measure this representative aspect of online budget allocation, we define the primary metric of \textit{first-sight commitment}, the fraction of commitments made in an episode on a class's first occurrence. Being conditional on committing, it excludes streams where the model never commits and can rate a rarely-building model as maximally eager, so we pair it with the outcome-based Score defined below and with three unconditional diagnostics defined in Appendix~\ref{app:bootstrap}.

Measuring when a model chooses to commit its budget does not provide a complete evaluation of budget allocation. To this end, we also equip the non-evaluative rungs of AllocBench (R0, R1, and R2) with an unbiased and comprehensive score metric, defined by \begin{equation}
\text{Score} = \frac{\sum_i \text{utility}_i}{\sum_i \text{optimal}_i}.
\label{eq:score}
\end{equation}
Here $\text{utility}_i$ is the total realized utility on episode $i$ (Equation~\ref{eq:utility}) and $\text{optimal}_i$ is the theoretical maximum realizable utility, defined by the sum of the sizes of the $B$ largest classes. A model's score is bounded above by 100\%, combining timing and allocation accuracy into one intuitive value. 

\section{Experimental Results}
\label{sec:results}

We evaluate five models in distinct roles. Claude Haiku 4.5 is the primary controlled subject; Claude Opus 4.8 provides a frontier paired test plus an R2 bridge; Qwen2.5-Coder-14B-Instruct is used for reward learning; GPT-5.4-mini and GPT-5.6 Sol are evaluated under a preregistered competence gate with matched R0 and R2 conditions. Because the models differ in reasoning settings, calibration, and hand-solving ability, the primary evidence throughout is paired, \emph{within-model} frame comparisons on identical streams; cross-model comparisons are secondary and not causally interpretable. Model settings, mechanical-validity checks, and rerun rules appear in Appendix~\ref{app:reconstruction}.

We organize our results in logical order. In Section~\ref{sec:gate}, we establish that 
frontier models naturally possess allocation ability in the abstract frame R0, but no model preserves this ability in the 
constructive environment R3. We isolate the failure mode for most models at code 
emission (R2) in Section~\ref{sec:ladder-results}. Next, Section~\ref{sec:score-results}
presents a simple numerical metric of online allocation ability unconfounded by varying tool 
creation ability. Finally, Section~\ref{sec:qwen-case-study} shows that the gap between R0 and R2 holds
even in models that have been post-trained specifically to succeed at R0. As a further robustness check, Appendix~\ref{app:economic} confirms that this gap is not an artifact of economic ambiguity;  abstract allocation responds to a changing visible build cost while the code-required policy does not.

\subsection{Abstract competence but tool-budgeting failure}
\label{sec:gate}

\begin{figure}
\centering
\includegraphics[width=0.7\linewidth]{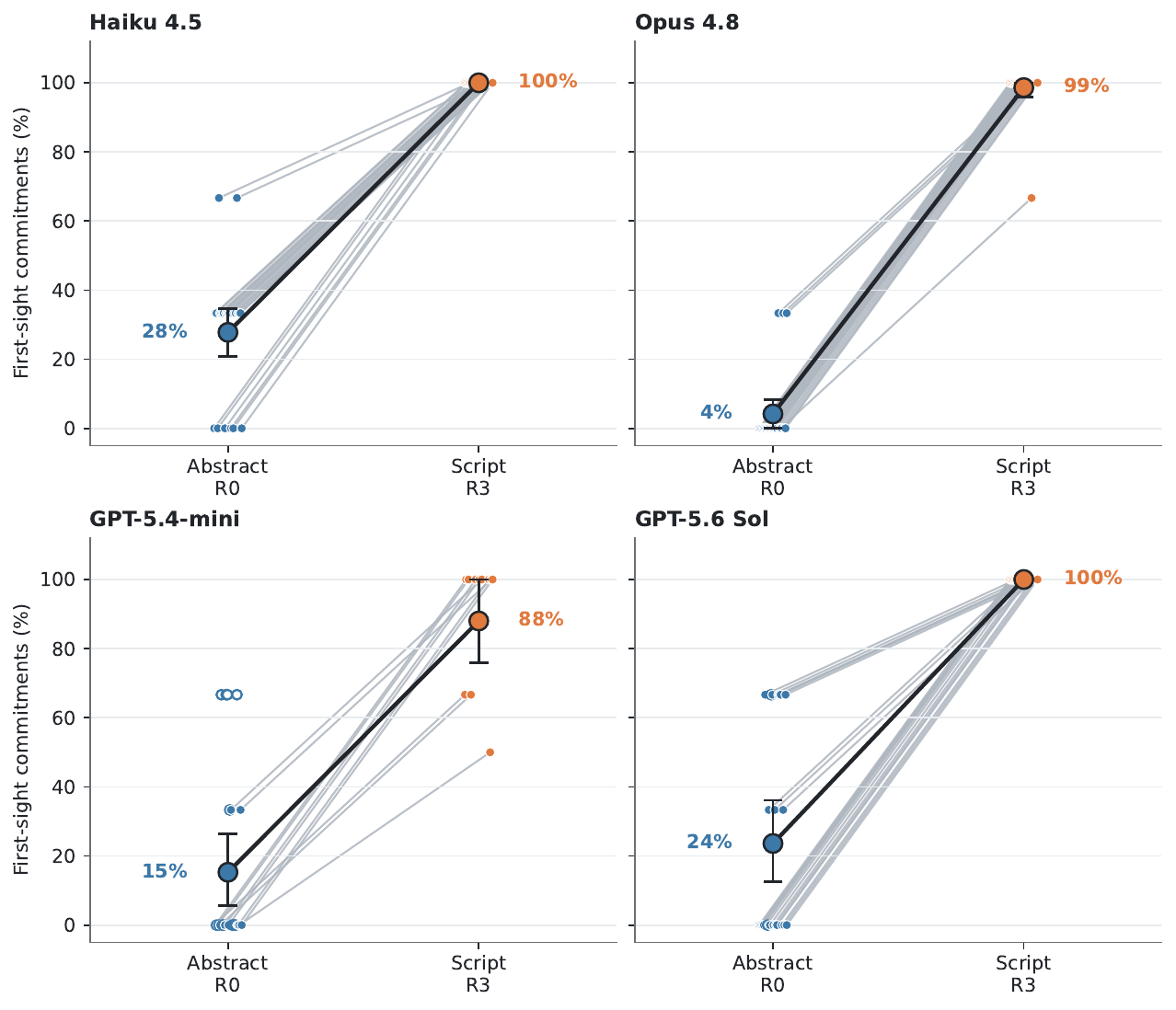}
\caption{\textbf{First-sight commitment rate of frontier models on R0 and R3.} Thin lines connect paired seeds; large points show averages, with 95\% seed-bootstrap intervals ($N=24$) on each pooled rate. Open circles mark seeds that commit in the
abstract frame but realize no script build at all in R3. The paired within-seed
R0$\to$R3 difference intervals are reported in
Appendix~\ref{app:paired-intervals}.}
\label{fig:core}
\end{figure}

We first show that frontier models have the ability to 
allocate budget semi-optimally in an abstract setting. We test Haiku, Opus, 
GPT-5.4-mini, and GPT-5.6 Sol on the R0 framing (see Section \ref{sec:frames}). We find that all four frontier models reserve their 
budgets: as shown in Figure \ref{fig:core}, each waits for evidence of recurrence before committing more than $70\%$ of the time. However, this competence fails to generalize to tool allocation. Though the
R0 and R3 environments provide the models with the same information and task them with the same 24 class streams, no model waits for recurrence more than $12\%$ of the time in the latter. This shows that R3 significantly suppresses a demonstrated allocation ability in all tested models.

The models' reasoning transcripts in the two frames help to explain the differences in their behavior. Abstract decisions prompt
references to recurrence and budget preservation: in one stream, Opus says, \textit{Still gathering information;
too early to commit a keep.} This is mirrored by the other models and reflects the intended behavior of committing only once a type's rate is
established (\textit{Red has now appeared 3 times in 7 draws --- a strong rate.
With 53 draws left, keeping red should yield many more.}) In contrast, the tool frame causes models to fixate on the immediate instance. For example, Opus states on the first problem of a stream, \textit{I'll write a script for this bit-manipulation
problem type.} Budget is only mentioned near exhaustion (\textit{I have 1 write left. Let me use it for this LCG
type which may recur.}) The models do not consider waiting for recurrence evidence at all, and scripts are reused only reactively when a class later returns. For fuller per-seed excerpts, see Appendix~\ref{app:transcripts}.

\subsection{The issue is code emission}
\label{sec:ladder-results}

In Section \ref{sec:frames}, we defined 4 framings of AllocBench problem streams that form a gradient from abstraction to construction. Table \ref{tab:ladder} reports pooled
first-sight commitment rates at each rung (see Appendix \ref{app:paired-intervals} for paired differences and confidence intervals). As established in Section \ref{sec:gate}, every model reserves budget in R0, committing at first sight less than $30\%$ of the time. Here we see that declarative claims on numeric problems (R1) do not cause a significant behavioral shift in any model, indicating that the failure is not induced by mathematical reasoning.\footnote{We note that the two smaller models, Haiku and GPT-5.4-mini, perform slightly worse on R1 than on R0, but the effect is small compared to the jump to R2.} However, the R2 rung, which does not check for code correctness but asks the model to generate a script for the problems it wishes to claim, practically eliminates any allocation ability of all models except GPT-5.6 Sol. Since Sol is able to maintain an R2 first-sight commitment rate comparable to that of Haiku on R1, allocation under code requirement is demonstrably possible for LLMs. However, even Sol fails to allocate its budget when forced to use the \texttt{write\_script} and \texttt{run\_script} tools. This points to a very strong bias that current models develop, whether in pre-training or post-training, that associates tool use with fixation on the current instance. 

\begin{table}
\centering
\small
\begin{tabular}{@{}|l|rrrr|c|@{}}
\hline
Model & R0 & R1 & R2 & R3 & Failure point \\
\hline
Haiku 4.5    & 28\%   & 33\% & \textbf{99\%}  & 100\%  & R2 \\
Opus 4.8     & 4\%    & 0\%  & \textbf{92\%} & 99\% & R2 \\
GPT-5.4-mini & 15\% & 29\% & \textbf{85\%} & 88\% & R2 \\
GPT-5.6 Sol  & 24\% & 0\%  & 39\% & \textbf{100\%}  & R3 \\
\hline
\end{tabular}
\caption{\textbf{Pooled first-sight commitment rates for each frontier model across the framing ladder} (Section \ref{sec:frames}). Only GPT-5.6 Sol preserves some allocation ability in R2, and all the models fail in R3.}
\label{tab:ladder}
\end{table}

\subsection{A comprehensive allocation metric}
\label{sec:score-results}

First-sight commitment rates, while indicative of model tendencies, provide a limited perspective of allocation capability. To make AllocBench generalizable for future research, we present our results using a simple score metric defined as the pooled competitive
ratio against the exact hindsight optimum (see Section \ref{sec:outcomes}). Note that we do not measure score on R3, because performance there reflects both allocation and script-writing capability. In the other rungs, score is an unconfounded measure of budget allocation.

\begin{figure}
    \centering
    \includegraphics[width=0.5\linewidth]{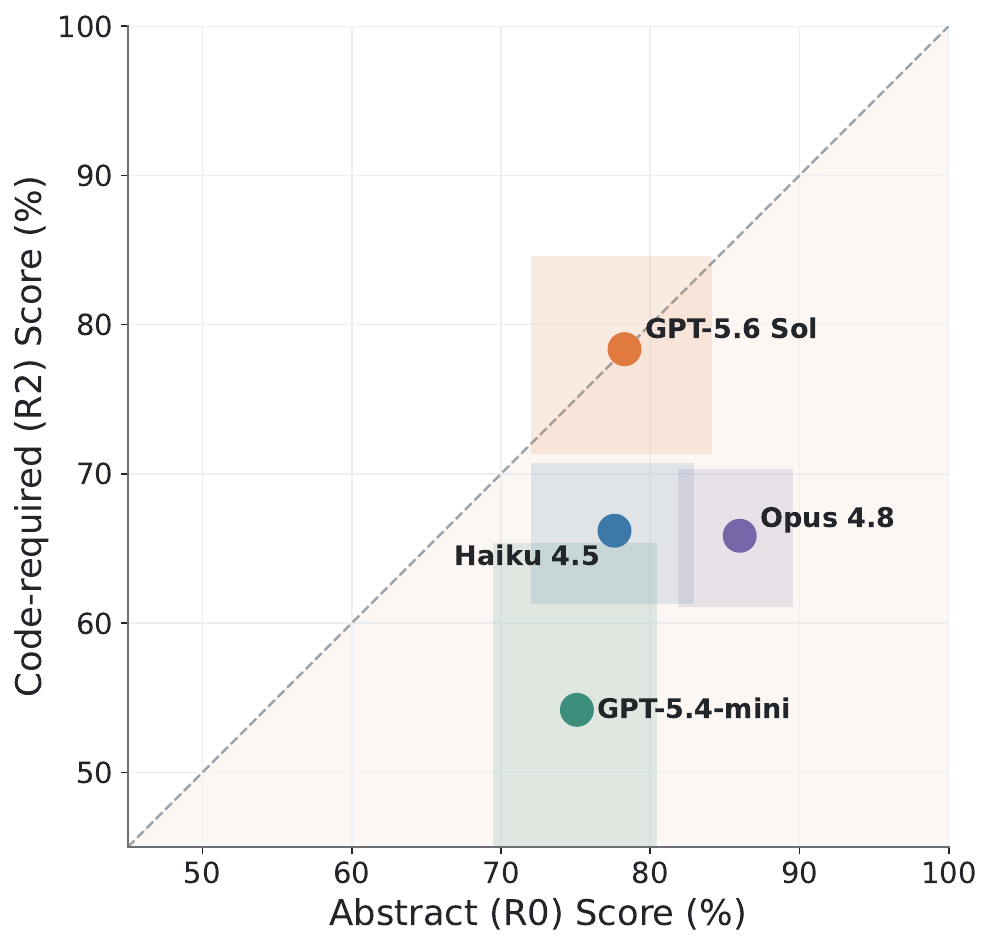}
    \caption{\textbf{Score (percent of the optimal net utility realized) for each frontier model in R0 vs. in R2.} The trend matches that in Table \ref{tab:ladder}. Note that only GPT-5.6 Sol avoids a performance drop from R0 to R2. Shaded boxes are 95\% seed-bootstrap intervals ($N=24$); they mark the marginal interval on each axis, not a joint confidence region (the R0 and R2 scores share seeds and are correlated).}
    \label{fig:score}
\end{figure}

Figure \ref{fig:score} reports the scores of the frontier models on R0 and R2. The results mirror the first-sight commitment statistics in Table \ref{tab:ladder}. The area highlighted in orange represents the region in which abstract allocation is easier than constructive allocation, which we assume must hold for any model. Points on the dotted line thus represent complete generalization of performance from R0 to R2, while points far from it indicate a failure to transfer abstract ability to the constructive setting. As shown, the four models all score within a $10\%$ interval on R0, but GPT-5.6 Sol scores nearly $25\%$ higher than GPT-5.4-mini on R2. The significant degradation of allocation ability, even among strong frontier models like Opus 4.8, is an intriguing and applicable direction for future model improvements.

\subsection{Learned abstract policies may not generalize}
\label{sec:qwen-case-study}

As an additional case study, we ask whether reinforcement learning can install an online allocation policy in a model that is initially eager in every frame, and how well such an acquired policy transfers. We start with the open-source Qwen2.5-Coder-14B-Instruct model, quantized at q8\_0 GGUF. We use Quantized Low-Rank Adaptation (QLoRA) \citep{dettmers2023qlora} with Proximal Policy Optimization (PPO) \citep{schulman2017ppo} to maximize the expected utility at every turn. Training uses no demonstrations; instead, we draw from robotics methodology \citep{pinto2018asymmetric} and employ an asymmetric critic that observes the true latent rate of the current class. More details can be found in Appendix \ref{app:rl}. 

\begin{figure}[H]
\centering
\includegraphics[width=0.9\linewidth]{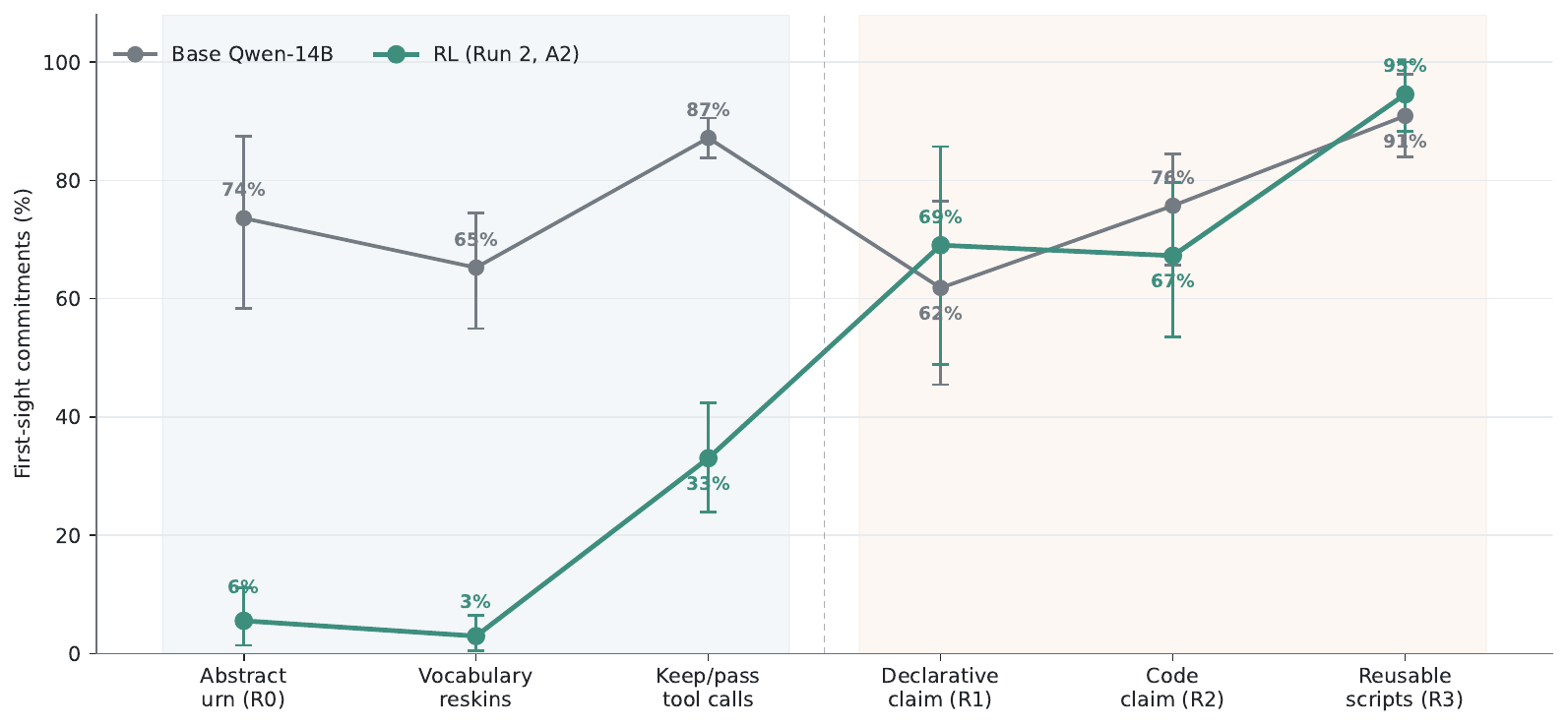}
\caption{\textbf{First-sight commitment rates for Qwen before and after post-training across the framing ladder} In the blue region, a learned abstract policy performs significantly better than the base model. Once a numerical problem is shown, the two models are statistically indistinguishable. Error bars are 95\% seed-bootstrap intervals ($N=24$) on each pooled rate. }
\label{fig:rl-transfer}
\end{figure}

We train for 20 steps at 25 episodes/step and find that R0 first-sight commitment, evaluated on the same held-out streams as the frontier models (Section \ref{sec:construction}), drops from $74\%$ to $6\%$ (see Figure \ref{fig:rl-transfer}). Since R0 streams all describe colored balls in an urn, we also test whether this learned behavior generalizes to lexical reskins, such as coins in a treasure chest or arrows in a quiver. We find that first-sight commitment on these reskins remains near-zero (in fact slightly lower than R0), indicating that the trained strategy is not limited to the ball framing. Further tests with the same abstract keep/pass choice, but issued through a required zero-argument tool call rather than free text, also show significant improvements.

Interestingly, transfer breaks down even at R1, before any frontier model's documented failure. Compared to the abstract keep/pass tool call framing, R1 only changes the setting from collecting items to solving math problems. This change makes the base and post-trained Qwens statistically indistinguishable under the metric of first-sight commitment (Figure \ref{fig:rl-transfer}). Computing the two versions' scores supports this picture; however, since score is a much narrower metric, the effects are less resolved (see Table \ref{tab:score}). The post-trained model reaches an R0 score comparable to a strong frontier model (79.6\%, versus GPT-5.6 Sol's 78.3\% from Figure \ref{fig:score}), while the base model performs worse than GPT-5.4-mini. In contrast, the acquired policy produces no score benefit once a numeric problem is shown.\footnote{In fact, the post-trained model performs worse at R2; this reflects a code-emission channel limitation of the 14B model rather than allocation timing (paired contrasts and channel diagnostics in Appendix~\ref{app:qwen-validity}). We therefore treat within-model first-sight commitment as the primary transfer signal.} This represents only one attempt at post-training online budget allocation, but the failure of a learned abstract policy to generalize to a numerical framing suggests that allocation suppression in R3 is not a quirk of any particular model architecture. We encourage future studies to apply existing budget-awareness methodologies \citep{liu2025budgetaware, liu2026intent, li2026bavt, lin2026bagen} to AllocBench and measure generalization on the framing ladder.  

\begin{table}
\centering
\begin{tabular}{|l|rrr|}
\hline
Model & R0 score & R1 score & R2 score \\
\hline
Qwen-14B base (q8)         & 74.8\% & 49.2\% & 61.2\%\\
Qwen-14B post-trained (q8) & 79.6\% & 34.8\% & 41.7\%\\
\hline
\end{tabular}
\caption{\textbf{Qwen Score on R0, R1, and R2, before and after post-training.}
Point estimates (percent of hindsight-optimal utility realized, pooled over the
24 evaluation seeds). Post-training leaves Score no higher at R1 or R2. Paired difference intervals and more details are given in Appendix~\ref{app:qwen-validity}.}
\label{tab:score}
\end{table}

\section{Discussion}
\label{sec:discussion}

\paragraph{Why might construction suppress allocation?}
We conjecture that required code emission invokes a strong post-trained bias that fixates
on solving the visible instance and takes precedence over
``invest for later'' considerations. This hypothesis is consistent with a characteristic of modern post-training: coding datasets such as DeepSWE invariably present streams of unrelated problems with independent rewards \citep{wei2025swerl, luo2025deepswe}. Because episodes share no state, a tool constructed while solving one
problem provides no expected future value. Furthermore, reasoning models are trained with explicit length and efficiency penalties \citep{team2025kimi}. A policy shaped this
way has no gradient toward the reserve behavior that R0 elicits and instead favors solving the visible instance as directly as possible---potentially resulting in the observed phenomenon of premature commitment \citep{chen2025greedy, mehta2026commit}. While frontier labs
do not disclose the exact training methods behind their coding models, both OpenAI and Anthropic have stated that they employ RL on coding datasets, making this a feasible explanation of allocation suppression \citep{openai2025competitive, anthropic2025claude37card}. We leave a complete mechanistic interpretation of this phenomenon for future work.

\paragraph{Implications for agent systems.}
LLM tool creation skills have been thoroughly studied, and script-writing agents are now ubiquitous. Our results indicate that current models cannot simultaneously manage budget allocation and tool creation. This suggests that tool-building agent systems should separate the abstract decision to invest from the process of creating the tool. For example, the R3 task streams could be split with one agent outputting declarative claims when building is optimal (R1), and a different agent constructing the desired tools. Since models preserve budget far better under R1 than R3, this simple separation could significantly boost performance on the latter. We hope that future work can test such joint agent systems, perhaps alongside methodologies that boost budget awareness \citep{liu2025budgetaware, lin2026bagen, liu2026intent, li2026bavt, wang2026mcpp}.

\paragraph{Implications for capability evaluation.} In this paper, we distinguish between two separate failure modes that existing tool benchmarks
conflate. A model can fail on a task because a policy is \textit{absent}, or because it is present but \textit{suppressed} by framing. For example, baseline Qwen possesses no allocation capability, but the frontier models we test have this ability suppressed by the constructive framing. Applying this framework to a wider range of tasks may yield a more detailed understanding of modern LLM capabilities. 

\section{Limitations}

\paragraph{Synthetic environment.} AllocBench uses one synthetic numeric environment with engineered classes and streams, designed to allow for precise measurements and comparisons. Real tasks have ambiguous class boundaries, shifting frequencies, incomplete reusability, and an unknown number of classes. Additionally, we conduct all experiments on 24 independent and randomly generated problem streams (see Appendix \ref{app:reconstruction}). While we include uncertainty estimates and our results are statistically significant, further experiments must be run to verify our results in different settings.

\paragraph{Reasoning off.} Both GPT models run at \texttt{reasoning\_effort=none} due to the OpenAI endpoint rejecting tools with reasoning on (see Appendix \ref{app:reconstruction} for more details). Since the paired design holds this fixed within model, it does not drive the R0$\to$R3 dissociation. Though it is possible that a larger reasoning budget might let these models recover some constructive-frame allocation, this is unlikely as the capability wall persists across a wide range of model capabilities.

\paragraph{RL case study.} Our RL experiments (Section \ref{sec:qwen-case-study}) are conducted with a privileged critic and serve only as an additional data point in favor of a capability wall between abstract and constructive framing. They do not demonstrate non-privileged
learnability, nor do they imply that it is impossible to post-train models to perform well even at script allocation. Similarly, we make no statements regarding the exact mechanistic cause of allocation suppression.

\paragraph{Relative scoring.} Our score metric (defined in Section \ref{sec:outcomes}) is calibrated by an exact hindsight optimum that is not implementable for an online model. Thus, score values should be referenced only as a measure of \textit{relative} model performance. As an alternative, we present first-sight commitment statistics to quantitatively measure a specific aspect of allocation behavior.

\section{Conclusion}

Budget-aware tool allocation is not only a coding problem; it is an online investment
decision under uncertain reuse. We introduce the first benchmark for budget allocation and run paired experiments to isolate the allocation capabilities of four frontier models and an open-source Qwen. We find that every frontier model demonstrates a conscious allocation policy in the abstract
frame R0 but fails to express it with script construction (R3). Even when Qwen is post-trained to behave optimally in R0, it fails to generalize to R3. Interestingly, required code emission alone (R2) suppresses all the models but GPT-5.6 Sol, suggesting that tool allocation is possible but very difficult for current models. We conjecture that this phenomenon is caused by existing coding datasets, which are commonly used for post-training, comprising many completely unrelated problems. We hope to see future work in two directions: explaining and resolving tool-based allocation suppression, and designing multi-agent systems that enable more efficient tool allocation in realistic tasks.

\section{Acknowledgments}

This work was substantially supported by Sea12 Technologies.

\section*{Reproducibility Statement}

We have taken several steps to make our results reproducible. The AllocBench benchmark formulation, stream generator, family set, and construction choices are introduced in Sections \ref{sec:decision-process} and \ref{sec:construction} and reconstructed in full detail in Appendix~\ref{app:reconstruction}, including verbatim example problem instances. The framing ladder is given in Section \ref{sec:frames}. All model settings, reasoning configurations, mechanical-validity checks, and rerun rules are documented in Appendix~\ref{app:reconstruction}; our statistical estimands and the seed-bootstrap inference procedure are defined in Appendix~\ref{app:bootstrap}. The reinforcement-learning acquisition and transfer study, including QLoRA/PPO hyperparameters and the asymmetric critic, is detailed in Appendix~\ref{app:rl}, and the exact commands and a manifest of the run artifacts underlying every figure and table are listed in Appendix~\ref{app:manifest}. Code, environments, and raw episode logs are available at \url{https://github.com/DragonWrangler25/AllocBench}.

\bibliography{allocation}

@article{cai2024latm,
  title   = {Large Language Models as Tool Makers},
  author  = {Cai, Tianle and Wang, Xuezhi and Ma, Tengyu and Chen, Xinyun and Zhou, Denny},
  journal = {ICLR},
  year    = {2024}
}

@article{qian2023creator,
  title   = {{CREATOR}: Tool Creation for Disentangling Abstract and Concrete Reasoning of Large Language Models},
  author  = {Qian, Cheng and Han, Chi and Fung, Yi R. and Qin, Yujia and Liu, Zhiyuan and Ji, Heng},
  journal = {Findings of EMNLP},
  year    = {2023}
}

@article{stengeleskin2024regal,
  title   = {{ReGAL}: Refactoring Programs to Discover Generalizable Abstractions},
  author  = {Stengel-Eskin, Elias and Prasad, Archiki and Bansal, Mohit},
  journal = {ICML},
  year    = {2024}
}

@article{grand2024lilo,
  title   = {{LILO}: Learning Interpretable Libraries by Compressing and Documenting Code},
  author  = {Grand, Gabriel and Wong, Lionel and Bowers, Matthew and Olausson, Theo X. and Liu, Muxin and Tenenbaum, Joshua B. and Andreas, Jacob},
  journal = {ICLR},
  year    = {2024}
}

@article{wang2024trove,
  title   = {{TroVE}: Inducing Verifiable and Efficient Toolboxes for Solving Programmatic Tasks},
  author  = {Wang, Zhiruo and Fried, Daniel and Neubig, Graham},
  journal = {ICML},
  year    = {2024}
}

@article{berlotattwell2024librarylearning,
  title   = {Library Learning Doesn't: The Curious Case of the Single-Use ``Library''},
  author  = {Berlot-Attwell, Ian and Rudzicz, Frank and Si, Xujie},
  journal = {MATH-AI Workshop, NeurIPS},
  year    = {2024}
}

@article{huang2024ultratool,
  title   = {Planning, Creation, Usage: Benchmarking {LLM}s for Comprehensive Tool Utilization in Real-World Complex Scenarios},
  author  = {Huang, Shijue and Zhong, Wanjun and Lu, Jianqiao and Zhu, Qi and Gao, Jiahui and Liu, Weiwen and Hou, Yutai and Zeng, Xingshan and Wang, Yasheng and Shang, Lifeng and Jiang, Xin and Xu, Ruifeng and Liu, Qun},
  journal = {Findings of ACL},
  year    = {2024}
}

@article{qu2025toollearningsurvey,
  title   = {Tool Learning with Large Language Models: A Survey},
  author  = {Qu, Changle and Dai, Sunhao and Wei, Xiaochi and Cai, Hengyi and Wang, Shuaiqiang and Yin, Dawei and Xu, Jun and Wen, Ji-Rong},
  journal = {Front. Comput. Sci.},
  year    = {2025}
}

@article{park2025regret,
  title   = {Do {LLM} Agents Have Regret? A Case Study in Online Learning and Games},
  author  = {Park, Chanwoo and Liu, Xiangyu and Ozdaglar, Asuman and Zhang, Kaiqing},
  journal = {ICLR},
  year    = {2025}
}

@article{chen2025greedy,
  title   = {When Greedy Wins: Emergent Exploitation Bias in Meta-Bandit {LLM} Training},
  author  = {Chen, Sanxing and Chen, Xiaoyin and Huang, Yukun and Xie, Roy and Dhingra, Bhuwan},
  journal = {arXiv preprint arXiv:2509.24923},
  year    = {2025}
}

@article{mehta2026commit,
  title   = {When Agents Commit Too Soon: Diagnosing Premature Commitment in {LLM} Agents},
  author  = {Mehta, Aman},
  journal = {arXiv preprint arXiv:2606.22936},
  year    = {2026}
}

@article{ding2026calibrate,
  title   = {Calibrate-Then-Act: Cost-Aware Exploration in {LLM} Agents},
  author  = {Ding, Wenxuan and Tomlin, Nicholas and Durrett, Greg},
  journal = {arXiv preprint arXiv:2602.16699},
  year    = {2026}
}

@inproceedings{wu2024counterfactual,
  title     = {Reasoning or Reciting? Exploring the Capabilities and Limitations of Language Models Through Counterfactual Tasks},
  author    = {Wu, Zhaofeng and Qiu, Linlu and Ross, Alexis and Aky{\"u}rek, Ekin and Chen, Boyuan and Wang, Bailin and Kim, Najoung and Andreas, Jacob and Kim, Yoon},
  booktitle = {Proceedings of the 2024 Conference of the North American Chapter of the Association for Computational Linguistics: Human Language Technologies (Volume 1: Long Papers)},
  year      = {2024}
}

@article{liu2025budgetaware,
  title   = {Budget-Aware Tool-Use Enables Effective Agent Scaling},
  author  = {Liu, Tengxiao and Wang, Zifeng and Miao, Jin and Hsu, I-Hung and Yan, Jun and Chen, Jiefeng and Han, Rujun and Xu, Fangyuan and Chen, Yanfei and Jiang, Ke and Daruki, Samira and Liang, Yi and Wang, William Yang and Pfister, Tomas and Lee, Chen-Yu},
  journal = {arXiv preprint arXiv:2511.17006},
  year    = {2025}
}

@article{lin2026bagen,
  title   = {{BAGEN}: Are {LLM} Agents Budget-Aware?},
  author  = {Lin, Yuxiang and Wang, Zihan and Liu, Mengyang and Shan, Yuxuan and Bai, Longju and Zhang, Junyao and Jin, Xing and Chen, Boshan and Su, Jinyan and Wang, Xingyao and Pei, Jiaxin and Li, Manling},
  journal = {arXiv preprint arXiv:2606.00198},
  year    = {2026}
}

@article{liu2026utility,
  title   = {Utility-Guided Agent Orchestration for Efficient {LLM} Tool Use},
  author  = {Liu, Boyan and Zhao, Gongming and Xu, Hongli},
  journal = {arXiv preprint arXiv:2603.19896},
  year    = {2026}
}

@article{li2026bavt,
  title   = {Spend Less, Reason Better: Budget-Aware Value Tree Search for {LLM} Agents},
  author  = {Li, Yushu and Deng, Wenlong and Li, Jiajin and Li, Xiaoxiao},
  journal = {arXiv preprint arXiv:2603.12634},
  year    = {2026}
}

@article{liu2026intent,
  title   = {Budget-Constrained Agentic Large Language Models: Intention-Based Planning for Costly Tool Use},
  author  = {Liu, Hanbing and Tian, Chunhao and An, Nan and Wang, Ziyuan and Lu, Pinyan and Yu, Changyuan and Qi, Qi},
  journal = {arXiv preprint arXiv:2602.11541},
  year    = {2026}
}

@article{wang2026mcpp,
  title   = {On Time, Within Budget: Constraint-Driven Online Resource Allocation for Agentic Workflows},
  author  = {Wang, Xinglin and Liu, Zishen and Feng, Shaoxiong and Yuan, Peiwen and Li, Yiwei and Shi, Jiayi and Zhang, Yueqi and Tan, Chuyi and Zhang, Ji and Pan, Boyuan and Hu, Yao and Li, Kan},
  journal = {arXiv preprint arXiv:2605.06110},
  year    = {2026}
}

@article{schulman2017ppo,
  title   = {Proximal Policy Optimization Algorithms},
  author  = {Schulman, John and Wolski, Filip and Dhariwal, Prafulla and Radford, Alec and Klimov, Oleg},
  journal = {arXiv preprint arXiv:1707.06347},
  year    = {2017}
}

@article{dettmers2023qlora,
  title   = {{QLoRA}: Efficient Finetuning of Quantized {LLM}s},
  author  = {Dettmers, Tim and Pagnoni, Artidoro and Holtzman, Ari and Zettlemoyer, Luke},
  journal = {NeurIPS},
  year    = {2023}
}

@article{pinto2018asymmetric,
  title   = {Asymmetric Actor Critic for Image-Based Robot Learning},
  author  = {Pinto, Lerrel and Andrychowicz, Marcin and Welinder, Peter and Zaremba, Wojciech and Abbeel, Pieter},
  journal = {Robotics: Science and Systems},
  year    = {2018}
}

@article{yu2026wildtoolbench,
  title   = {Benchmarking LLM Tool-Use in the Wild},
  author  = {Yu, Peijie and Liu, Wei and Yang, Yifan and Li, Jinjian and Zhang, Zelong and Feng, Xiao and Zhang, Feng},
  journal = {ICLR},
  year    = {2026}
}

@article{wei2025swerl,
  author = {Wei, Yuxiang and Duchenne, Olivier and Copet, Jade and Carbonneaux, Quentin and Zhang, Lingming and Fried, Daniel and Synnaeve, Gabriel and Singh, Rishabh and Wang, Sida I.},
  title = {{SWE-RL}: Advancing {LLM} Reasoning via Reinforcement Learning on Open Software Evolution},
  journal = {NeurIPS},
  year = {2025}
}

@misc{luo2025deepswe,
  author = {Luo, Michael and Jain, Naman and Singh, Jaskirat and Tan, Sijun and Patel, Ameen and Wu, Qingyang and Ariyak, Alpay and Cai, Colin and Venkat, Tarun and Zhu, Shang and Athiwaratkun, Ben and Roongta, Manan and Zhang, Ce and Li, Li Erran and Popa, Raluca Ada and Sen, Koushik and Stoica, Ion},
  title = {{DeepSWE}: Training a Fully Open-Sourced, State-of-the-Art Coding Agent by Scaling {RL}},
  year = {2025},
  howpublished = {\url{https://www.together.ai/blog/deepswe}},
  note = {Agentica and Together AI. Accessed 2026-07-21}
}

@article{team2025kimi,
  author = {{Kimi Team}},
  title = {{Kimi k1.5}: Scaling Reinforcement Learning with {LLM}s},
  journal = {arXiv preprint arXiv:2501.12599},
  year = {2025}
}

@article{openai2025competitive,
  author = {{OpenAI}},
  title = {Competitive Programming with Large Reasoning Models},
  journal = {arXiv preprint arXiv:2502.06807},
  year = {2025}
}

@misc{anthropic2025claude37card,
  author = {{Anthropic}},
  title = {{Claude 3.7 Sonnet} System Card},
  year = {2025},
  howpublished = {\url{https://www.anthropic.com/claude-3-7-sonnet-system-card}},
  note = {Accessed 2026-07-21}
}

\newpage

\appendix

\section{Benchmark Theory and Preregistration} \label{app:design-details}

\subsection{Optimal-policy arithmetic} \label{app:policy-math}

The prompts disclose only that $N=8$ classes occur; the rates and the three-hot/five-trap split are never revealed (Appendix~\ref{app:reconstruction}), so we price decisions under a symmetric Dirichlet($\alpha$) prior over the eight rates, with $\alpha=1$ (uniform on the simplex) as the canonical choice. A class that has appeared $k$ times in $t$ draws has marginal rate posterior $\mathrm{Beta}(\alpha+k,\,7\alpha+t-k)$ and expected future occurrences $(T-t)(\alpha+k)/(8\alpha+t)$ -- about $13$ at a draw-1 first sight and $14$ at a draw-4 second occurrence. The posterior mean alone barely separates these two cases; what recurrence moves is the class's \emph{rank}. With $B=3$ slots the decision-relevant event is membership in the top three rates: at first sight $P(\text{top-}3)$ is at most $(3H_8-\tfrac{5}{2})/8\approx 0.71$ -- the prior expected top-three mass, attained when the first sight is draw 1 and lower for later first sights -- whereas a second occurrence within the first half-dozen draws gives $0.74$--$0.88$ (Monte Carlo, $2{\times}10^{6}$ posterior draws per configuration). The errors are priced asymmetrically: deferring forfeits the current occurrence (one point) and slightly shortens the reuse horizon, while committing spends one of three slots on a class outside the top three with probability at least $0.29$. In prior expectation these nearly cancel: over $4{\times}10^{4}$ simulated streams, commit-at-second-occurrence scores $37.0$ points versus $36.7$ at first sight and $35.5$ at third, an ordering unchanged at $\alpha\in\{0.5,2\}$, so deferring costs nothing even for an agent with no design knowledge. On the benchmark's realized distribution -- bimodal rates ($r_h = 0.28\overline{3}$, $r_\tau = 0.03$) with a trap forced into the first $B$ draws -- the preference becomes strict and large: $43.0$ versus $34.6$ points per stream. For reference, an observer who knew the generative recipe (three hot, five trap, both rates, but not the assignment) would hold posterior odds $\tfrac{3}{5}\,(r_h/r_\tau)^{k}\,\big(\tfrac{1-r_h}{1-r_\tau}\big)^{t-k}$: $P(\text{hot})$ is exactly $0.85$ at a draw-1 first sight (at most $0.85$ in general) and $0.94$--$0.97$ at a typical second occurrence, with a slot spent on a trap forfeiting a hot class's ${\approx}15$-point reuse stream. This is a strictly stronger information state than the agent's; both lead to the same defer-then-commit rule.

\subsection{Preregistration and localization criterion} \label{app:prereg}

The ladder's steps are composite (R0$\to$R1 changes modality, labels, and content together; R1$\to$R2 adds required code emission and its attendant instructions and schema; R2$\to$R3 adds execution, persistence, and correctness-dependent consequences), so it identifies the earliest composite rung that shifts behavior, not a single causal factor. The preregistered rule declares the trigger at the earliest adjacent increase of at least 30 percentage points in first-sight commitment (pooled among realized commitments, to which zero-commit seeds contribute nothing; the unconditional hazard is reported alongside) -- a minimum effect of interest, roughly half the piloted R0-to-R3 gap and about four seed-bootstrap standard errors at $N=24$. All other ladder comparisons are descriptive and not multiplicity-adjusted. The hypotheses, rule, and analysis plan were frozen before data collection. Prompt and harness repairs are categorized as pre- or post-outcome in the artifact record.

\section{Experimental Reconstruction} \label{app:reconstruction}

\subsection{Streams and session state}

All confirmatory Haiku, Opus, and GPT comparisons use canonical seeds 2000--2023. For each seed, a local pseudorandom generator first permutes the eight procedure families \texttt{lcg}, \texttt{modpow}, \texttt{continued\_frac}, \texttt{crt\_solve}, \texttt{josephus}, \texttt{quadratic\_map\_mod}, \texttt{xorshift\_steps}, and \texttt{matrix\_power\_mod} across three hot and five trap roles. Hot classes jointly receive probability 0.85 and traps 0.15, uniformly within role. The generator samples $T=60$ class identities i.i.d. and rejection-samples until at least one trap occurs in the first $B$ positions ($B=3$ in the core comparison). Numeric instances are generated at magnitude 100 for the canonical Haiku, GPT, and primary Opus R3 conditions. The Opus R3 arm holds the complete latent stream fixed---family, class ID, role, rate, realized count, occurrence position, and arrival order---and reuses the same magnitude-100 numeric rendering as Haiku and both GPT models, so all four models answer literally identical problem instances. It does not pin or relocate any family. Structural projections match the canonical Opus urn artifacts on 24/24 seeds. Class metadata are hidden; only rendered item or problem text is shown to the model. Haiku R0, R1, R2, and R3 reuse the same saved magnitude-100 stream for each seed. Appendix~\ref{app:magnitude-sensitivity} separately reports a magnitude-1000 completeness check for Opus's R3 arm, the pool used earlier in this project to foreclose Opus's partial hand-solving ability.

Each commitment consumes one of $B$ slots and is irreversible. Subsequent members of a committed class are automatically collected or solved and produce no new decision. Sessions stop making decisions when the budget is exhausted, but scoring includes all later members of committed classes. Economic sessions use the same streams and cross $B\in\{1,3,5\}$ with $K\in\{0,10,20,24\}$; $K$ changes only the disclosed one-time charge and net score.

\subsection{Prompt and tool reconstruction}

The artifact retains every rendered user message, assistant response, tool call, and generated stream. Exact prompt generators and schemas are in \texttt{urn\_common.py}, \texttt{urn\_session.py}, \texttt{urn\_tool\_session.py}, \texttt{claim\_solver\_session.py}, \texttt{claim\_solver\_code\_session.py}, and \texttt{session\_state.py}. To remove ambiguity about substitutions, the core prompts have $T=60$, $N=8$, and the current $B$ inserted into the following templates.

\paragraph{R0 allocation instructions.} The system states that 60 items are drawn one at a time; at any draw the model may KEEP the current item or PASS; at most $B$ items may be kept; keeping an item immediately collects it and every future item with the same visible attribute; passing scores zero; already-kept attributes are auto-collected and do not create another decision; and choices do not change the stream. The disclosure is:
\begin{quote}\small
There are exactly 8 distinct colors in the bag, but you are not told which colors are common or rare, nor how many balls of each color there are; you learn the rates only from the draws you have seen.
\end{quote}
R0 ends with the exact response instruction
\begin{quote}\small
For each draw, think briefly (one or two sentences), then end your reply with EXACTLY one line: \texttt{DECISION: KEEP} or \texttt{DECISION: PASS}.
\end{quote}

\paragraph{R1/R2 allocation instructions.} Both prompts state that 60 numeric problems arrive one at a time, each has a hidden type, exactly eight types occur, rates are initially unknown, a solver may be claimed at most $B$ times, a claim scores the current and all future problems of that type, skipping scores zero, decisions are final, and the model must call exactly one of \texttt{claim\_solver} or \texttt{skip\_solver}. R1 explicitly says not to compute an answer and uses two zero-argument tools. R2 instead adds the following exact requirement:
\begin{quote}\small
You must provide \texttt{code}: a real, correct Python function \texttt{def solve(inputs: dict) -> float} that solves problems of this type---it stands in for every future problem of the same type for the rest of the session, so it should actually work, not just handle the problem currently shown.
\end{quote}
The R2 \texttt{claim\_solver} schema therefore has one required string property, \texttt{code}; \texttt{skip\_solver} remains a zero-argument object. The prompt does not disclose that correctness is excluded from reward. For an economic cell, both R0 and R2 add the exact statement that committing costs $K$ points once, each collected or solved problem remains worth one point, and declining costs nothing. R3 uses the reusable-script harness's \texttt{write\_script}, \texttt{run\_script}, and answer tools; the complete prompt and schema generators are in \texttt{driver.py} and \texttt{session\_state.py}, and complete resulting transcripts are stored in \path{runs/arm_a1_announce_n-announced/} for Haiku and \path{runs/arm_a1_announce_opus_n-announced_canonical-structure/} for the repaired Opus endpoint.

\subsection{Example rendered instances} \label{app:example-instances}

The generators above are stated only by name in the main text; this subsection shows problems exactly as the model sees them. All items are quoted verbatim from the canonical magnitude-100 streams (\path{stream.json}), including the \texttt{xorshift\_steps} item that opens the seed-2000 session whose Opus rationale is quoted in Appendix~\ref{app:transcripts}. Three families are shown, one magnitude-hard and two structure-hard; hidden class metadata (family, role, recurrence rate) is never rendered.

{\small
\begin{verbatim}
xorshift_steps  (seed 2000, problem 1; gold = 1098337718)
  A 32-bit register holds X = 32586213. Repeat 14 times:
  X ^= (X << 11); X ^= (X >> 9); X ^= (X << 17); keeping X masked
  to 32 bits (AND with 0xFFFFFFFF) after each left shift.
  Report the final X.

crt_solve  (gold = 112394190734)
  Solve the system of congruences: x is congruent to
  [320, 1069, 449, 1959] modulo [557, 1553, 461, 2003] respectively
  (the moduli are pairwise coprime). Give the least non-negative x.

josephus  (gold = 87)
  168 people stand in a circle numbered 1..168. Counting around the
  circle, every 6-th person is eliminated (continuing past those
  already gone) until one remains. Report the position of the
  survivor.
\end{verbatim}
}

To make the randomized-cover-story mechanism concrete, the following three \texttt{lcg} instances---drawn from different seeds---each wrap the identical recurrence $X \leftarrow (aX + c) \bmod m$ in an unrelated surface narrative. Recognizing that they share one procedure, and that a single script solves all of them, therefore requires operation-level recognition rather than string-matching on the prompt.

{\small
\begin{verbatim}
lcg, cover A  (gold = 2537700)
  Initialize s = 141. Repeat the following 26 times:
  s = (73100 * s + 38400) % 21759900. Give the resulting value
  of s.

lcg, cover B  (gold = 15476700)
  A pseudorandom counter begins at 103. For 23 rounds,
  set X = (59900 * X + 67400) mod 23773700. What is X after the
  final round?

lcg, cover C  (gold = 14634500)
  A register starts at X = 272. It is updated 28 times; each
  update replaces X with ((43300 * X) + 38700) mod 18314600.
  Report the final value of X.
\end{verbatim}
}

\subsection{Models and generation settings}

Claude runs use \texttt{claude-haiku-4-5-20251001} and \texttt{claude-opus-4-8}. GPT runs use \texttt{gpt-5.4-mini-2026-03-17} and \texttt{gpt-5.6-sol}. Provider-default temperature is held fixed across paired arms. Claude R1 and R2 use \texttt{tool\_choice=auto}; forcing a tool call suppressed deliberation in calibration. GPT uses \texttt{reasoning\_effort=none} in both arms and \texttt{tool\_choice=auto} for R2. Maximum output is 512 tokens for R0, 1500 for R1, 2500 for R2, and 4096 per R3 turn, with a 300,000-token session cap. Opus R3 sessions stop immediately after the third successful script write; all 24 terminate on budget exhaustion without hitting the token or per-seed cost breaker, and contain no malformed or unknown calls. GPT-5.4-mini R3 sessions truncate on write-budget exhaustion when any script is written (6/24 seeds) and otherwise run to the 300,000-token session cap or full stream completion (18/24 seeds); all 24 are mechanically clean, with zero malformed, unknown, or refused tool calls. GPT-5.6 Sol R3 sessions likewise truncate on budget exhaustion for 12/24 seeds and on the 300,000-token session cap for the other 12; all 24 are mechanically clean. Qwen starts from \texttt{Qwen/Qwen2.5-Coder-14B-Instruct}; matched final tool evaluations use q8\_0 base and adapted models with an 8192-token context.

\paragraph{Reasoning effort and the paired design.} Because both GPT models run at \texttt{reasoning\_effort=none} in every arm, one might ask whether their constructive-frame failures reflect a genuine allocation deficit or merely a throttled reasoning budget. This setting was not a free parameter for the tool rungs in our harness: the \texttt{/v1/chat/completions} endpoint we use rejects function-tool schemas with reasoning on for both GPT models (a 400 directing callers to \texttt{/v1/responses}), so any rung that attaches tools (R1--R3) runs at \texttt{reasoning\_effort=none}. We hold R0 at \texttt{none} as well for within-model parity. The evidence is insulated from this by construction: the identical setting is used in both arms of every within-model comparison, so each GPT model demonstrates near-optimal R0 allocation at the very reasoning budget under which it collapses at R2/R3. The R0$\to$R3 dissociation is therefore not attributable to reasoning effort, which is held fixed within model. Two further observations argue against a reasoning-configuration artifact: GPT-5.6 Sol retains partial allocation ability at R2 under the same setting, and the Claude models---which emit deliberation text at every rung---fail at R2/R3 just as sharply.

\subsection{Competence gates}

A GPT model passes the locked abstract competence gate only if all 24 R0 sessions complete, it makes at least 2.5 commitments per seed, and no more than 50\% of realized commitments are first-sight.\footnote{The locked preregistration includes a fourth criterion---mean R0 collection at least 90\% of a same-information online Bayesian (Dirichlet) comparator on the identical streams. We omit it from the description above because it proved non-binding: both gated models satisfy the three criteria stated here independently and clear the comparator bar with margin (GPT-5.4-mini 0.96, GPT-5.6 Sol 1.00), so dropping it changes neither gate outcome. Its symmetric-Dirichlet prior is misspecified for the engineered generator, and we do not report the comparator elsewhere; the locked analysis code still computes and applies it.} The preregistered suppression replication additionally requires at least 80\% R2 first-sight among commitments, a pooled R2--R0 increase of at least 30 points, and a positive lower endpoint for the paired bootstrap interval (met by GPT-5.4-mini; GPT-5.6 Sol instead satisfies it at R3, consistent with its suppression appearing one rung later).

\subsection{Mechanical validity and rerun rules} \label{app:mech-validity}

Transport or provider exceptions receive at most two additional attempts with linear 1.5-second backoff. Model-level failures---no decision, multiple decisive calls, an unknown function, malformed arguments, or an unparsable response---are not retried; they default to the non-commit action and remain counted as unresolved. A ladder rung is invalid if unresolved decisions exceed 10\%. Response-model identity, stream hash, malformed and unknown calls, termination reason, usage, and token-cap hits are checked separately from behavior. Only a mechanical failure permits a replacement run. For the final Qwen script evaluation, empty-fence retries are fixed at four in both arms and all attempts remain logged.

\paragraph{Cross-type recognition in class-bound R3.} The class-bound R3 harness binds each written script to the hidden class of the instance it was written on and refuses any attempt to run it on a later instance of a different class, logging the refusal in \texttt{n\_cross\_class\_run\_refused} without executing the code or charging the write budget. This count is a direct, if partial, probe of type recognition: a refusal marks a case where the model invoked a script whose bound class does not match the instance in front of it. Across the magnitude-100 class-bound R3 sessions, 22 of 478 run attempts (4.6\%) were refused as type-mismatched, and the rate was lowest for the strongest models (Opus 1/66; GPT-5.6 Sol 0/133; GPT-5.4-mini 5/63) and higher for the 14B Qwen arms (base-q8 8/74). The metric captures only false-positive reuse---running the wrong script---not the converse failure of not recognizing that two instances share a type; and because eager first-sight commitment exhausts the three-write budget early, the reuse window over which this is observed is truncated. With those caveats, observed cross-type misapplication is rare, consistent with the recognition component of the joint bound in Section~\ref{sec:frames} being cheap rather than the binding constraint.

\section{Statistical Estimands and Inference} \label{app:bootstrap}

\subsection{Conditional and unconditional outcome measures}

A turn is \emph{eligible} iff $c_t \notin M_t$, $b_t > 0$, and the model returned a valid, non-truncated response (invalid turns are excluded as mechanical failures). Let $\mathrm{occ}_c(t) = |\{t' \le t : c_{t'} = c\}|$; the \emph{lateness} of a commitment at turn $t$ is $\mathrm{occ}_{c_t}(t) - 1$ (within-class occurrences, not stream positions: committing at a class's second occurrence has lateness~1), reported descriptively. Three further seed-level measures, unlike the conditional first-sight rate, retain zero-commit seeds. \textbf{First-sight hazard} is the committed share of eligible first-occurrence turns, pooled over eligible turns rather than over realized commitments; it is the primary timing view in the economic surface (Appendix~\ref{app:economic}). \textbf{Budget utilization} is realized commitments divided by the budget $B$, and \textbf{zero-commit incidence} is the share of seeds with no commitment; both are reported per frontier model and rung in Appendix~\ref{app:zero-commit} (Table~\ref{tab:zero-commit}) and accompany the reward-learning ladder (Table~\ref{tab:rl-transfer-full}).

\subsection{Paired seed-bootstrap procedure}

For each paired comparison, first-sight among commitments is computed within seed, the second-arm value minus the first-arm value is taken, and paired seeds are sampled with replacement 10,000 times using bootstrap RNG seed 0. The reported point estimate is the arithmetic mean of seed differences and the interval is the percentile interval. A seed with no commitments has an undefined among-commitments proportion and is omitted only from that conditional estimand; it remains in commitment, zero-commit, hazard, and utility summaries.

\subsection{Re-pooled-ratio estimand}

Two paired tables instead report the closely related \emph{difference of re-pooled ratios} estimand -- the cross-model Table~\ref{tab:frontier-paired} and the cross-checkpoint Table~\ref{tab:rl-transfer-full} (and its main-text figure, Figure~\ref{fig:rl-transfer}) -- which resamples the same 24 seeds as clusters, re-pools each arm's among-commitment ratio within a resample, and differences those pooled ratios; this variant retains zero-commit seeds (they add nothing to either pooled ratio) rather than dropping them from a conditional per-seed mean, and the two estimands coincide where every seed commits. Each table's caption states which it uses.

\subsection{Economic endpoint interactions}

Economic endpoint interactions are computed per seed as \[ \{h_{R0}(0)-h_{R0}(24)\} -\{h_{R2}(0)-h_{R2}(24)\}, \] separately for each $B$, and then resampled by seed. No session within a seed is treated as an independent replicate.

\section{Supplementary Frontier-Model Results} \label{app:supp-results}

\subsection{Paired framing-ladder intervals} \label{app:paired-intervals}

The main-text figures carry per-arm 95\% intervals on each rung's pooled first-sight rate, which are the quantities a level plot can display. The inferential claims of Section~\ref{sec:gate} (the R0$\to$R3 dissociation of Figure~\ref{fig:core}) and Section~\ref{sec:ladder-results} (the localization of Table~\ref{tab:ladder}) are instead \emph{paired within-seed differences}, which a level plot cannot show; Table~\ref{tab:frontier-paired} collects them. Each entry is a paired seed bootstrap of the difference of the two rungs' re-pooled first-sight-among-commitments ratios, resampling the 24 canonical seeds (2000--2023) as clusters 10,000 times at RNG seed 0 -- the same estimand and procedure as Table~\ref{tab:rl-transfer-full}. Zero-commit seeds are retained (they contribute nothing to either re-pooled ratio). The overall R3$-$R0 column corresponds to the paired dissociation plotted per model in Figure~\ref{fig:core}; the adjacent columns localize it along the ladder and recover the preregistered required-code trigger (the large, interval-excluding-zero jump at R2$-$R1 for Haiku, Opus, and GPT-5.4-mini; at R3$-$R2 for GPT-5.6 Sol).

\begin{table}[H]
\centering \caption{\textbf{Paired within-seed first-sight commitment differences for each frontier model}, in percentage points with 95\% seed-bootstrap intervals ($N=24$, 10,000 resamples, RNG seed 0; difference of re-pooled among-commitment ratios, matching Table~\ref{tab:rl-transfer-full}). Adjacent-rung columns localize the shift; the final column is the overall R0$\to$R3 dissociation of Figure~\ref{fig:core}. Positive means more first-sight (eager) commitment at the later rung.} \label{tab:frontier-paired} \small \resizebox{\linewidth}{!}{%
\begin{tabular}{@{}lcccc@{}}
\hline
Model & R1$-$R0 & R2$-$R1 & R3$-$R2 & R3$-$R0 (overall) \\
\hline
Haiku 4.5    & $+5.6$ $[-11.1,+23.6]$  & $+65.3$ $[+51.4,+79.2]$ & $+1.4$ $[+0.0,+4.2]$    & $+72.2$ $[+65.3,+79.2]$ \\
Opus 4.8     & $-4.2$ $[-8.3,+0.0]$    & $+91.7$ $[+86.1,+97.2]$ & $+6.9$ $[+0.0,+13.9]$   & $+94.4$ $[+88.9,+98.6]$ \\
GPT-5.4-mini & $+14.1$ $[-10.8,+40.5]$ & $+55.6$ $[+31.2,+77.1]$ & $+3.0$ $[-14.3,+22.2]$  & $+72.7$ $[+57.8,+86.2]$ \\
GPT-5.6 Sol  & $-23.6$ $[-36.1,-12.5]$ & $+38.9$ $[+25.0,+52.8]$ & $+61.1$ $[+47.2,+75.0]$ & $+76.4$ $[+63.9,+87.5]$ \\
\hline
\end{tabular}%
}
\end{table}

\subsection{Per-rung marginal intervals}

Table~\ref{tab:ladder} reports each ladder cell as a point estimate; Table~\ref{tab:frontier-cell} gives the same cells with 95\% seed-bootstrap intervals on the pooled first-sight rate. These are \emph{per-arm marginal} intervals: because the design is paired within seed, they do not settle whether two rungs differ -- overlapping level intervals can accompany a difference whose own interval excludes zero, and vice versa -- so the significance of any rung-to-rung shift is read from the paired differences of Table~\ref{tab:frontier-paired}, not from overlap here. Point estimates match Table~\ref{tab:ladder} (which rounds to integers); intervals degenerate to a point where a rung sits at the $0\%$ or $100\%$ ceiling on every seed.

\begin{table}[H]
\centering \caption{\textbf{Per-cell first-sight commitment for each frontier model across the ladder}, pooled among realized commitments, with 95\% seed-bootstrap intervals ($N=24$ canonical seeds, 10,000 resamples, RNG seed 0; per-arm marginal intervals, re-pooled ratio). These annotate the point estimates of Table~\ref{tab:ladder}; rung-to-rung significance is given by the paired differences of Table~\ref{tab:frontier-paired}, not by interval overlap here.} \label{tab:frontier-cell} \small \resizebox{\linewidth}{!}{%
\begin{tabular}{@{}lcccc@{}}
\hline
Model & R0 & R1 & R2 & R3 \\
\hline
Haiku 4.5    & $27.8$ $[20.8,34.7]$ & $33.3$ $[19.4,48.6]$ & $98.6$ $[95.8,100.0]$ & $100.0$ $[100.0,100.0]$ \\
Opus 4.8     & $4.2$ $[0.0,8.3]$    & $0.0$ $[0.0,0.0]$    & $91.7$ $[86.1,97.2]$    & $98.6$ $[95.8,100.0]$ \\
GPT-5.4-mini & $15.3$ $[5.6,26.4]$  & $29.4$ $[7.7,54.5]$  & $85.0$ $[70.2,96.7]$    & $88.0$ $[76.0,100.0]$ \\
GPT-5.6 Sol  & $23.6$ $[12.5,36.1]$ & $0.0$ $[0.0,0.0]$    & $38.9$ $[25.0,52.8]$    & $100.0$ $[100.0,100.0]$ \\
\hline
\end{tabular}%
}
\end{table}

\subsection{Zero-commit incidence and commitment volume} \label{app:zero-commit}

Table~\ref{tab:zero-commit} reports, for each frontier model and rung, the \textbf{zero-commit incidence} (seeds with no commitment) and the mean number of commitments per seed, over the 24 canonical seeds (2000--2023). Commitments are counted by each rung's native convention: \texttt{kept} classes at R0, \texttt{claimed} classes at R1--R2, and authored scripts at R3 (so a zero-commit R3 seed authors no script at all). Zero-commit incidence is nil for every model except the two competence-gated GPT models under required code. GPT-5.6~Sol spends the full budget through R2 and drops only two seeds at R3. GPT-5.4-mini is the outlier: it already under-commits at the declarative rung R1 (17/24 zero-commit, 0.71 commits per seed), recovers most of its budget at R2 (4/24, 2.50), and collapses again under full construction at R3 (13/24, 1.04). The paired R2$\to$R3 changes (10{,}000-resample seed bootstrap, RNG seed~0; Appendix~\ref{app:bootstrap}) are, for GPT-5.4-mini, a $+37.5$-point $[+12.5,+62.5]$ rise in zero-commit incidence and a $-1.46$ $[-2.13,-0.75]$ fall in commits per seed, and for Sol, $+8.3$ points $[+0.0,+20.8]$ and $-0.79$ $[-1.21,-0.42]$. Haiku and Opus retain essentially the full budget at every rung, so their allocation failures are a matter of \emph{timing} rather than of withholding commitment.

\begin{table}[H]
\centering \caption{\textbf{Zero-commit incidence (seeds with no commitment, out of 24) and mean commitments per seed for each frontier model across the ladder.} Canonical seeds 2000--2023, all cells at $N=24$. R0/R1/R2 count committed classes (\texttt{kept}/\texttt{claimed}); R3 counts authored scripts, so a zero-commit R3 seed authors none. Zero-commit is negligible except for the competence-gated GPT models under required code (R1--R3).} \label{tab:zero-commit} \small
\begin{tabular}{@{}lcccccccc@{}}
\hline
 & \multicolumn{4}{c}{Zero-commit ($n/24$)} & \multicolumn{4}{c}{Mean commits / seed} \\
\cline{2-5}\cline{6-9}
Model & R0 & R1 & R2 & R3 & R0 & R1 & R2 & R3 \\
\hline
Haiku 4.5    & 0 & 0  & 0 & 0  & 3.00 & 3.00 & 3.00 & 2.83 \\
Opus 4.8     & 0 & 0  & 0 & 0  & 3.00 & 3.00 & 3.00 & 3.00 \\
GPT-5.4-mini & 0 & 17 & 4 & 13 & 3.00 & 0.71 & 2.50 & 1.04 \\
GPT-5.6 Sol  & 0 & 0  & 0 & 2  & 3.00 & 3.00 & 3.00 & 2.21 \\
\hline
\end{tabular}
\end{table}

\subsection{Magnitude sensitivity check} \label{app:magnitude-sensitivity}

The main text reports Opus's R3 arm at magnitude 100, matching Haiku and both GPT models so that all four models in the panel answer literally the same problem instances. This magnitude was not the arm originally used earlier in this project: an initial calibration found that at magnitude 100, Opus hand-solves roughly 25--38\% of problems in isolation (\texttt{josephus} alone at essentially 100\%), so a magnitude-1000 rendering was adopted specifically to foreclose that escape valve before the R3 timing result was trusted. We include that magnitude-1000 arm here purely for completeness, not as the primary result.

Rerunning the identical 24 canonical seeds, harness, and truncation protocol at magnitude 1000, Opus makes 72/72 R3 builds at first sight (100\%) with zero mean lateness; the R3-minus-R0 paired difference is $+95.8$ points, 95\% seed-bootstrap interval $[+91.7,+100.0]$. This is statistically indistinguishable from the magnitude-100 result reported in the main text (98.6\% first sight, 71/72 builds, mean lateness 0.014, paired difference $+94.4$ points $[+88.9,+98.6]$): the two intervals overlap substantially, and neither timing estimate is degenerate at $N=24$. Josephus accounts for seven of the 72 builds in both arms, all at first sight in both, so its known hand-solvability at any magnitude does not drive either result.

The magnitude change therefore has no detectable effect on Opus's construction-time eagerness. We report the magnitude-100 arm as primary because it removes an asymmetry across the model panel, not because the magnitude-1000 arm turned out to be confounded: at magnitude 100, with a genuine if narrow hand-solve alternative available on every turn, Opus submitted a raw, script-free answer to an unbuilt class only once across the full 24-seed panel, and that attempt was incorrect. Artifacts are in \path{runs/arm_a1_announce_opus_n-announced_canonical-structure_mag100_class-bound-v1} (magnitude 100, this appendix's comparison arm and the main text's primary result) and \path{runs/arm_a1_announce_opus_n-announced_canonical-structure_class-bound-v1} (magnitude 1000, the completeness check).

\section{Economic Sensitivity: Inelasticity to Build Price} \label{app:economic}

\paragraph{Purpose and scope.} The framing ladder (Section~\ref{sec:ladder-results}) localizes the capability wall to required code emission at a single, fixed price ($K=0$). This leaves one counterpoint open: perhaps R2's near-ceiling eagerness does not reflect an inability to allocate under code emission, but merely a benchmark run at a price where committing immediately happens to be cheap. The comprehensive-Score result of Section~\ref{sec:score-results} already refutes the strongest form of this objection -- R2 realizes strictly less utility than R0 at $K=0$, so eager building is not harmless -- but leaves a weaker \emph{calibration} version standing: that the model \emph{could} allocate under code if the price signal were loud enough. This appendix closes that version. The study is a single-model robustness check on Claude Haiku ($N=24$ seeds, 2000--2023); we do not claim cross-model generalization. \textbf{Its result is that R2's construction-time eagerness is inelastic to build price: raising the visible charge makes abstract allocation (R0) progressively more selective, tracking the hindsight optimum, while the code-required policy (R2) keeps building at essentially the same rate even in the near-never-build regime, where the optimum builds almost nothing.} Price calibration does not recover the allocation ability that code emission suppresses.

\paragraph{Design.} We hold the latent class stream fixed and manipulate only the visible build charge $K$ and budget $B$, crossing abstract R0 and code-required R2 with $B\in\{1,3,5\}$ and $K\in\{0,10,20,24\}$. The charges span $K=0$, where eager construction of hot classes can pay; $K=10$, where hot classes pay but traps do not; and $K=24$, a near-never-build regime in which the exact hindsight optimum commits on only about 1\% of eligible first-sight turns. On the $N=24$ panel the $K=0,20,24$ grid contains 432 sessions and the added $K=10$ selective grid 144; all pass stream, parsing, and token-cap checks.

\begin{figure}[t]
\centering \includegraphics[width=\linewidth]{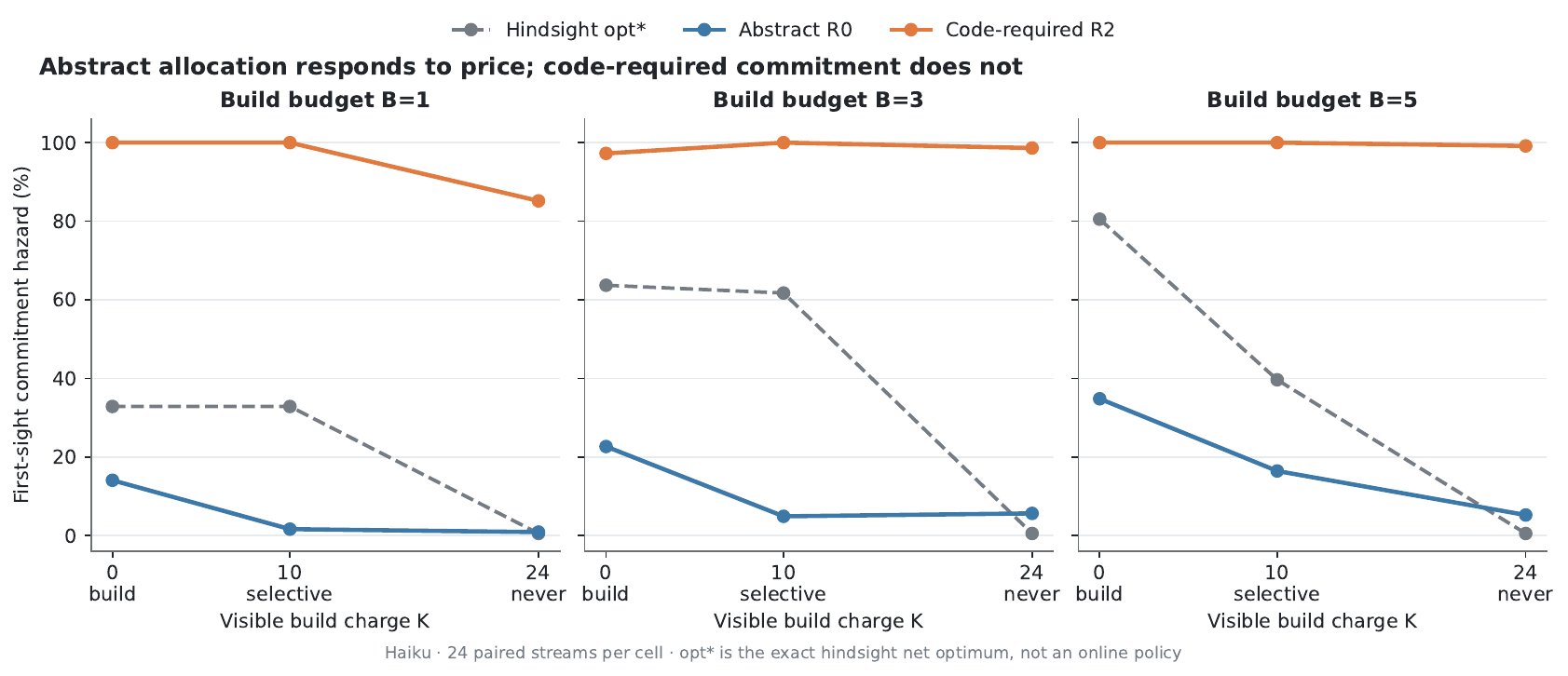} \caption{\textbf{Abstract allocation responds to price; code-required commitment does not.} First-sight commitment hazard across visible build charges for each build budget. The abstract R0 policy becomes more selective as construction grows expensive, broadly co-moving with the exact hindsight net optimum, which itself declines to near-zero commitment at $K=24$. The code-required R2 policy stays near 100\% first sight across the entire surface, including at $K=24$ where the optimum commits on only about 1\% of eligible first-sight turns -- the behavioral inelasticity whose cost appears as the rising R2 regret in Table~\ref{tab:econ-full}. Each cell reuses the same 24 paired Haiku streams; opt* is an offline upper bound, not an online policy.} \label{fig:economic}
\end{figure}

\paragraph{Utility quantifies the cost; timing shows the behavior.} The two views play complementary roles here. The \emph{optimum's} price response lives almost entirely in \emph{how many} classes clear the profitability bar, not \emph{when} they are built: the hindsight optimum builds 3.00 classes per seed at $K=0$ and 0.04 at $K=24$ (at $B=3$), while its build-lateness stays 0 throughout. A timing metric therefore cannot serve as the \emph{quantitative} price-response outcome, and the competitive-ratio Score of Section~\ref{sec:score-results} degenerates where the optimum's utility approaches zero (its denominator $\to 0$ at high $K$, driving Score to large negative values). We accordingly report \textbf{net utility and regret against the exact hindsight optimum} as the primary quantitative outcomes (Table~\ref{tab:econ-full}). The \emph{model's} inelasticity, however, is most legible as a behavioral signature -- whether its commitment timing moves with price at all -- so we use first-sight hazard for that: R0's timing tracks the falling optimum while R2's does not budge (Figure~\ref{fig:economic}).

Table~\ref{tab:econ-full} reports these quantities on the plotted charge surface, $N=24$ seeds (2000--2023). Hazard is pooled over eligible first-sight turns; all other columns are means over the 24 seeds. Zero-commit incidence is 0\% in every displayed cell.

\begin{table}[H]
\centering \scriptsize \caption{Economic surface: first-sight hazard and net-utility outcomes reported together. Net is collected or solved items minus $K$ times commitments; Optimum is the exact hindsight net optimum (frame-independent); Regret is Optimum minus Net.} \label{tab:econ-full} \resizebox{0.7\linewidth}{!}{%
\begin{tabular}{llrrrr}
\hline
Frame & $(B,K)$ & Hazard (\%) & Net & Optimum & Regret \\
\hline
R0 & $(1,0)$  & 14.1 & 10.6 & 19.5 & 8.9 \\
R0 & $(1,10)$ & 1.7  & 3.1 & 9.5 & 6.4 \\
R0 & $(1,24)$ & 0.9  & -10.5 & 0.1 & 10.6 \\
R0 & $(3,0)$  & 22.7 & 37.5 & 49.7 & 12.1 \\
R0 & $(3,10)$ & 4.9  & 10.1 & 19.7 & 9.6 \\
R0 & $(3,24)$ & 5.7  & -33.4 & 0.1 & 33.5 \\
R0 & $(5,0)$  & 34.8 & 51.0 & 55.9 & 4.9 \\
R0 & $(5,10)$ & 16.5 & 1.6 & 19.7 & 18.1 \\
R0 & $(5,24)$ & 5.3  & -61.8 & 0.1 & 61.8 \\
\hline
R2 & $(1,0)$  & 100.0 & 10.0 & 19.5 & 9.5 \\
R2 & $(1,10)$ & 100.0 & 0.0 & 9.5 & 9.5 \\
R2 & $(1,24)$ & 85.2 & -14.1 & 0.1 & 14.2 \\
R2 & $(3,0)$  & 97.3 & 32.7 & 49.7 & 17.0 \\
R2 & $(3,10)$ & 100.0 & 2.9 & 19.7 & 16.8 \\
R2 & $(3,24)$ & 98.6 & -38.5 & 0.1 & 38.6 \\
R2 & $(5,0)$  & 100.0 & 51.5 & 55.9 & 4.4 \\
R2 & $(5,10)$ & 100.0 & 1.5 & 19.7 & 18.2 \\
R2 & $(5,24)$ & 99.2 & -68.5 & 0.1 & 68.6 \\
\hline
\end{tabular}%
}
\end{table}

The utility columns carry the result. As $K$ rises, R0 grows more selective -- its commitments thin out and its net utility declines in the same direction as the hindsight optimum (imperfectly: R0 still overspends somewhat at high charges) -- whereas R2 spends its full budget at every charge, including $K=24$, where the optimum builds almost nothing. R2's regret therefore overtakes R0's as price rises: at $B=5$ the two are comparable at $K=0$ (regret 4.4 vs 4.9) but diverge to $68.6\pm1.2$ (R2) versus $61.8\pm4.5$ (R0) at $K=24$. The secondary timing view agrees: R0's first-sight hazard falls from 14--35\% at $K=0$ to 1--6\% at $K=24$ and its mean lateness rises with price, while R2 holds at 85--100\% first sight across the surface. We summarize the elasticity with a seed-paired endpoint interaction -- the $K{=}0$-to-$24$ reduction in first-sight hazard under R0 minus the corresponding reduction under R2: the mean interaction is $+30.8$ points $[+6.7,+54.2]$ for $B=1$, $+20.2$ $[+10.3,+30.3]$ for $B=3$, and $+29.9$ $[+18.5,+42.0]$ for $B=5$ (95\% seed-bootstrap intervals, $N=24$). Because R2 is near ceiling throughout, the interaction is driven almost entirely by R0's price response. Figure~\ref{fig:economic} shows this behavioral signature directly: R0's first-sight hazard descends with price alongside the optimum's, while R2 stays pinned near ceiling across the whole charge axis -- the flat R2 line is the inelasticity that the widening regret above quantifies.

\section{Qwen Training and Transfer Evaluation} \label{app:rl}

\subsection{Training protocol and optimization}

Both analyzed runs use QLoRA and per-decision PPO for 20 outer steps, 25 fresh training seeds per step, four rollouts per seed, one epoch per update, learning rate $6\times10^{-5}$, and rollout temperature 1.2. Training seeds begin at 9000 and do not overlap evaluation seeds. The critic observes the actor's ordinary state plus true class-rate features; the actor never receives those rates. Exact Monte Carlo returns are used without discounting. No demonstrations enter training. The runs differ in one respect: the first run's training rollouts predate the $N$-disclosure protocol and do not disclose $N$; the second run's training rollouts disclose $N$, matching the evaluation condition. Both runs' step-0 credit-assignment gates pass (positive mean advantage for hot-class keeps and trap passes, negative for trap keeps and first-sight hot passes), and both produce monotone reserve acquisition over training. Each reported checkpoint is the last of a step budget fixed in advance, taken with no early stopping and no evaluation-based checkpoint selection.

\subsection{Run nomenclature and pilot population}

Throughout, \textbf{Run~1} denotes the No-$N$ training variant and \textbf{Run~2} the checkpoint of the main text. Table~\ref{tab:rl-transfer-full} gives the full first-sight ladder for all three checkpoints with both paired contrasts against base.

\subsection{Training curves}

\begin{figure}[H]
\centering \includegraphics[width=0.95\linewidth]{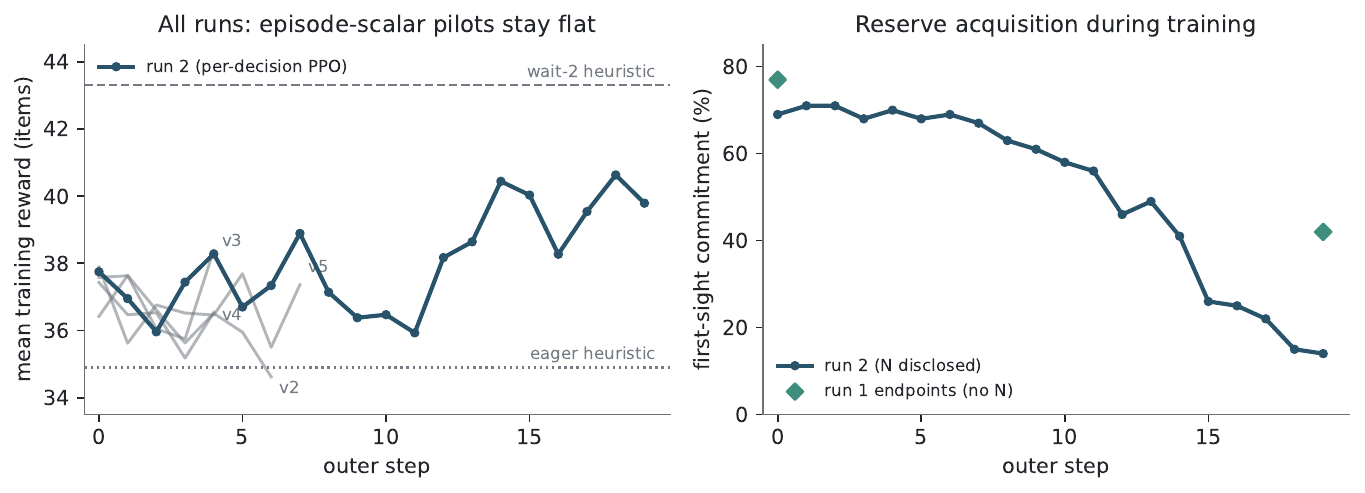} \caption{\textbf{Training curves for the complete run population.} Left: mean training reward (items collected) per outer step for the four flat episode-scalar GRPO pilots (gray) and the successful per-decision PPO Run~2 (navy), against the mean rewards of the eager and wait-for-one-repeat heuristics on the same training batches. Right: Run~2's per-step first-sight commitment rate on its own training rollouts, falling monotonically from 69\% to 14\% over 20 updates. Run~1's per-step history was not archived; its logged endpoints (77\%$\to$42\% on training batches) are overlaid as diamonds. Per-step training-batch metrics are computed on-policy at sampling temperature 1.2 and therefore sit above the corresponding held-out, temperature-0 evaluation numbers.} \label{fig:rl-training}
\end{figure}

\subsection{Transfer evaluation}

\begin{table}[H]
\centering \caption{\textbf{Full first-sight commitment ladder for all three Qwen checkpoints}, pooled over realized commitments (parentheses: mean lateness, descriptive; keep counts in the occurrence rows). Paired seed-bootstrap 95\% intervals versus base resample the 24 evaluation seeds as clusters and difference the re-pooled ratios. Run~1 is the No-$N$ training variant; Run~2 is the checkpoint of the main text (Figure~\ref{fig:rl-transfer}). Negative $\Delta$ = more reserve; items/stream higher is better. All panels: seeds 2000--2023, disclosed $N$, q8\_0 serving; R1--R3 use four empty-fence retries. $^\dagger$Exploratory: unresolved turns default to PASS, inflating apparent reserve (Appendix~\ref{app:qwen-validity}). $^\ddagger$Descriptive only: risk sets are policy-induced, so no paired interval is given.} \label{tab:rl-transfer-full} \resizebox{\textwidth}{!}{%
\begin{tabular}{lccccc}
\hline
Panel (rung) & Base & Run 1 (No-$N$) & $\Delta_1$ [95\% CI] & Run 2 ($N$-disclosed) & $\Delta_2$ [95\% CI] \\
\hline
Abstract urn (R0) & 74\% (0.40) & 35\% (0.75) & $-38.9$ $[-58.3,-19.4]$ & 6\% (1.51) & $-68.1$ $[-81.9,-51.4]$ \\
Lexical reskins (pooled) & 65\% (0.59) & 29\% (0.84) & $-36.1$ $[-48.0,-23.2]$ & 3\% (1.46) & $-62.3$ $[-71.5,-52.0]$ \\
Keep/pass tool call$^\dagger$ & 87\% (0.36) & 50\% (1.15) & $-37.2$ $[-46.2,-29.0]$ & 33\% (1.56) & $-54.1$ $[-64.1,-44.2]$ \\
Declarative claim (R1) & 62\% (1.49) & 57\% (1.68) & $-4.7$ $[-28.3,+19.1]$ & 69\% (0.95) & $+7.2$ $[-17.0,+30.0]$ \\
Code claim (R2) & 76\% (1.24) & 73\% (1.03) & $-2.7$ $[-16.0,+10.8]$ & 67\% (0.91) & $-8.4$ $[-22.8,+6.1]$ \\
Script (R3, class-bound) & 91\% (0.11) & 94\% (0.06) & $+3.4$ $[-5.2,+12.0]$ & 95\% (0.06) & $+3.6$ $[-6.0,+12.7]$ \\
\hline
P(keep), occurrence 1 (R0)$^\ddagger$ & 58\% (53/91) & 23\% (25/109) & -- & 3\% (4/131) & -- \\
P(keep), occurrence 2 (R0)$^\ddagger$ & 57\% (12/21) & 83\% (40/48) & -- & 60\% (48/80) & -- \\
Items/stream (R0) & 37.2 & 39.9 & $+2.8$ $[-1.8,+7.1]$ & 39.5 & $+2.4$ $[-2.7,+7.5]$ \\
\hline
\end{tabular}}
\end{table}
Per-vocabulary reskin first-sight rates span 57--79\% (base), 18--36\% (Run~1), and 1--6\% (Run~2), and every vocabulary's paired interval individually excludes zero (worst case Run~1 quiver, $\Delta_1=-23.6$ $[-44.4,-2.8]$; Run~2 ranges from $-50.9$ to $-77.0$). Conditioning keep decisions on occurrence count (Table~\ref{tab:rl-transfer-full}, occurrence rows) confirms a defer-then-confirm profile rather than a uniformly lower keep rate: the base keeps at a flat rate across the first two occurrences, whereas Run~2 almost never keeps on a first sighting and usually keeps on the second. These rates are descriptive only, as the risk sets are policy-induced (Run~2 sees more second occurrences because it passed on the first).

Each unchanged final checkpoint is evaluated across the full ladder: the abstract urn on the original ball vocabulary and on treasure-chest/coin/crest, quiver/arrow/feather, and cauldron/potion/label reskins (all retaining $N=8,T=60,B=3$, the same latent streams, disclosed $N$, and free-text keep/pass decisions); a probe that changes only the decision channel to required zero-argument keep/pass calls; and the declarative-claim (R1), code-required claim (R2), and reusable-script (R3) rungs of the main framing ladder. Both trained checkpoints acquire reserve on the abstract urn and generalize it across reskins and the tool-call channel, and both lose it at R1 and beyond --- Run~2 more sharply throughout, consistent with its N-disclosed training matching the evaluation protocol. Tool-call transfer is uniform across vocabularies for Run~1 (49--51\% first sight, versus 83--93\% for base) and wider-spread for Run~2 (24--43\%, pooled 33\%); an earlier probe panel had shown strong vocabulary sensitivity, which did not reproduce under the final quantization-matched protocol.

\subsection{Channel validity and retry diagnostics} \label{app:qwen-validity}

Channel validity varies by rung: unresolved decision turns (no tool call emitted; defaulting to the non-commit action) are absent on the free-text urn and reskins, reach 66/72/76\% (base/Run~1/Run~2) on the keep/pass tool-call probe, remain at 18/31/15\% at R1 and 46/64/73\% at R2 after four retries, and are fully eliminated by retries at R3; the remainder of this subsection discusses the resulting bias directions. At R1 the arms also differ in claim volume (claims/seed 2.29/2.33/1.75; zero-claim seeds 3/3/7 of 24).

\paragraph{Qwen R1/R2 exceed the 10\% threshold; we report them anyway, unfiltered.} All three Qwen R2 arms exceed the panel's 10\% unresolved-decision threshold by a wide margin -- base-q8 resolves 186/347 decisions (46.4\% unresolved), RL-final 146/402 (63.7\%), and Run~2 187/692 (73.0\%) -- and the R1 arms exceed it more modestly (17.5\%, 30.5\%, and 15.0\% unresolved), against zero malformed or unknown calls for every other model in the Score panel (Figure~\ref{fig:score}). This is the same idle-tail empty-fence generation failure common to base and the trained runs, not an RL-training artifact, and it survives the same \texttt{empty-fence-retry=4} mitigation used for the final Qwen script evaluation (Appendix~\ref{app:mech-validity}): the retry breaks runaway repetition and lets most sessions still spend their full write budget (22/24 base-q8, 17/24 RL-final, and 14/24 Run-2 seeds place all 3 claims), but does not restore a clean channel. Unresolved turns default to the non-commit action, so this pathology can depress commitment counts and manufacture apparent reserve -- it cannot manufacture the first-sight eagerness these arms show among realized claims, which is why we treat the R1/R2 nulls as conservative while flagging the keep/pass tool-call probe (66--76\% of turns unresolved, same default) as exploratory. Rather than invoke the mechanical-failure exception to drop or resample the affected seeds, we report Score pooled over all 24 seeds for every arm, unresolved decisions included, because the arms fail at different rates (2/24 base-q8 versus 7/24 RL-final and 10/24 Run-2 seeds never place their third claim before the stream ends): excluding a model's own failed seeds would remove more of precisely the outcomes that scored worst for whichever arm fails more often, manufacturing a base-vs-RL gap out of the exclusion rule itself rather than out of behavior. The unfiltered numbers (Score 61.2\% for base and RL-final, collected 729 and 730 of 1192; 41.7\% for Run~2, collected 497) are therefore the conservative, selection-bias-free comparison, at the cost of reflecting a genuine 14B capability limitation on this harness's code-writing tool schema rather than a fully clean channel. We disclose this explicitly as a departure from the 10\% rule rather than silently reporting numbers the stated methodology would otherwise call invalid.

Table~\ref{tab:qwen-score-paired} gives the paired base-vs-post Score contrasts that qualify the point estimates of Table~\ref{tab:score}. No rung shows a significant training gain: the R0 contrast is unresolved (the R0 acquisition is carried by first-sight timing, not Score; Table~\ref{tab:rl-transfer-full}), R1 straddles zero, and the only resolved effect is the R2 decrease, which reflects the code-emission channel limitation quantified above rather than allocation timing.

\begin{table}
\centering
\begin{tabular}{@{}lrrr@{}}
\hline
Rung & Base & Post-trained & $\Delta$ (post $-$ base) [95\% CI] \\
\hline
R0 (abstract urn)  & 74.8\% & 79.6\% & $+4.8$ $[-5.6,+14.4]$ \\
R1 (declarative)   & 49.2\% & 34.8\% & $-14.4$ $[-28.8,+0.3]$ \\
R2 (code)          & 61.2\% & 41.7\% & $-19.5$ $[-31.1,-7.4]$ \\
\hline
\end{tabular}
\caption{\textbf{Paired Score contrasts, post-trained (Run~2) minus base.} Score
is percent of hindsight-optimal utility realized; $\Delta$ is the difference of
re-pooled ratios with a 95\% seed-bootstrap interval (24 seeds 2000--2023
resampled as clusters, 10{,}000 resamples, RNG seed~0), the same estimand as
Table~\ref{tab:rl-transfer-full}. R1/R2 use four empty-fence retries. The R0
interval includes zero (Score is insensitive to the timing shift that first-sight
captures), R1 includes zero, and only R2 excludes zero -- a decrease attributable
to the unresolved-turn channel limitation (Run~2 leaves 73\% of R2 decision turns
unresolved) rather than to allocation.}
\label{tab:qwen-score-paired}
\end{table}

\subsection{R3 commitment-volume diagnostics}

The R1/R2/R3 rungs use four empty-fence retries in every arm, matching the main framing ladder. The reusable-script comparison uses 24 paired seeds 2000--2023 under the class-bound-r3-v1 protocol with matched q8\_0 quantization and calibrated script success; write-budget slots are attributed to the hidden class they can execute on, repairing the allocation-unit mismatch with R2. For base/Run~1/Run~2, R3 first-sight builds are 50/55, 50/53, 52/55; distinct-class commitments per seed 2.29/2.21/2.29; zero-build seeds 0/1/0; unconditional first-sight hazards 33.3/34.0/39.1\% --- the near-identical first-sight rates are not an artifact of differing denominators. All three R3 arms are mechanically clean after retries (zero malformed, unknown, or refused tool calls across all 72 sessions); the base arm truncates on write-budget exhaustion for 20/24 seeds and the 300,000-token session cap for the remaining 4/24, Run~1 on budget exhaustion for 23/24 and the token cap for 1/24, and Run~2 likewise across its 24 sessions.

\section{Transcript Excerpts} \label{app:transcripts}

This appendix collects verbatim model rationales behind the qualitative claim in Section~\ref{sec:gate}: abstract (R0) decisions reason prospectively about recurrence and budget, whereas full-construction (R3) decisions center on the instance in front of the model, register budget only near exhaustion, and reuse scripts reactively. All excerpts are quoted verbatim (ellipses mark elision) and are drawn from the same models and canonical streams (seeds 2000--2023) reported in the main text. Provenance is given as run directory and seed; per-seed transcripts contain the full turn-by-turn context.

\paragraph{R0: prospective reserve and confirmation on recurrence.} In the abstract urn frame, commitments are deferred explicitly to gather frequency evidence and are triggered only once a type's rate is established.
\begin{itemize}
\item Haiku, \path{runs/urn_haiku_n-announced}, seed 2000, third draw (\textsc{pass}): ``I've seen three different colors in three draws (red, blue, green). With 60 draws total and 8 colors in the bag, I still need more information about the frequency distribution. Passing to gather more data is the right strategy at this early stage.''
\item Opus, \path{runs/urn_opus_n-announced}, seed 2000, second draw (\textsc{pass}): ``Still gathering information; too early to commit a keep.''
\item Opus, \path{runs/urn_opus_n-announced}, seed 2000, seventh draw (\textsc{keep}, third occurrence of the type): ``Red has now appeared 3 times in 7 draws --- a strong rate. With 53 draws left, keeping red should yield many more.''
\end{itemize}

\paragraph{R3: instance-focused construction, late budget awareness, reactive reuse.} In the full-construction frame the same models write a script for the problem currently on screen. Budget is mentioned only when nearly gone, and reuse is considered only after a type recurs rather than reserved for in advance.
\begin{itemize}
\item Opus, \path{runs/arm_a1_announce_opus_n-announced_canonical-structure_mag100_class-bound-v1}, seed 2000, problem 1 (first write, no budget reasoning): ``I'll write a script for this bit-manipulation problem type.'' (The problem in front of the model here is the \texttt{xorshift\_steps} item shown in full in Appendix~\ref{app:example-instances}.)
\item Opus (same run), seed 2008, first write: ``I'll write a CRT solver script since this type may recur.'' Then, on its third and last write: ``I have 1 write left. Let me use it for this LCG type which may recur.'' --- the first mention of the write budget in the session arrives only as it is exhausted.
\item GPT-5.4-mini, \path{runs/arm_a1_announce_gpt-5.4-mini-2026-03-17_n-announced_write-budget-v2_class-bound-v1}, seed 2000, reuse decision: ``I can reuse the existing script only if the hidden type matches; here I'll just calculate directly since it's simple.''
\end{itemize}

GPT-5.6 Sol is omitted from the excerpts above because its R3 sessions emit no visible deliberation text (its snapshot runs at \texttt{reasoning\_effort=none}); its construction decisions are observable only through the tool-call sequence, which shows the same front-loaded, first-sight pattern.

\section{Commands and Artifact Manifest} \label{app:manifest}

The core runs are reproduced from the repository root with the $N=24$ seed panel (2000--2023): {\scriptsize
\begin{verbatim}
PYTHONPATH=. python -u -m scripts.ladder.urn_session \
  --model haiku --announce-n --seeds $(seq 2000 2023)
PYTHONPATH=. python -u -m scripts.ladder.urn_tool_session \
  --model haiku --announce-n --seeds $(seq 2000 2023)
PYTHONPATH=. python -u -m scripts.ladder.claim_solver_session \
  --model haiku --announce-n --seeds $(seq 2000 2023)
PYTHONPATH=. python -u -m scripts.ladder.claim_solver_code_session \
  --model haiku --announce-n --seeds $(seq 2000 2023)
PYTHONPATH=. python -u -m scripts.ladder.arm_a1_announce \
  --model haiku --announce-n --full-stream --seeds $(seq 2000 2023) --cap 20.0 --unit-cap 2.0
PYTHONPATH=. python -u -m scripts.ladder.arm_a1_announce \
  --model opus --announce-n --canonical-structure --magnitude 100 --publication \
  --seeds $(seq 2000 2023) --cap 3.0 --unit-cap 0.75
PYTHONPATH=. python -u -m scripts.economic.run_economic_surface \
  --seeds $(seq 2000 2023)
PYTHONPATH=. python -u -m scripts.analysis.analyze_economic_surface \
  --seeds $(seq 2000 2023) --charges 0 10 20 24
\end{verbatim}
} Core artifacts are \path{runs/urn_haiku_n-announced}, \path{runs/urn_tool_haiku_n-announced}, \path{runs/claim_solver_haiku_n-announced}, \path{runs/claim_solver_code_haiku_n-announced}, \path{runs/arm_a1_announce_n-announced}, and \path{runs/economic_surface_haiku}. The primary Opus R3 artifact (magnitude 100) is \path{runs/arm_a1_announce_opus_n-announced_canonical-structure_mag100_class-bound-v1}; the magnitude-1000 completeness check (Appendix~\ref{app:magnitude-sensitivity}) is \path{runs/arm_a1_announce_opus_n-announced_canonical-structure_class-bound-v1}. GPT artifacts are under \path{runs/economic_surface_gpt-5.4-mini-2026-03-17} and \path{runs/economic_surface_gpt-5.6-sol}; each contains the exact model identifier, settings, stream, transcript, usage, and per-seed analysis.

\section{Use of Large Language Models} \label{app:llms}

The authors used Claude Code and Cursor to conduct literature reviews and produce Python scripts to execute experiments. Additionally, LLMs were used occassionally to suggest further experiments. The
authors provided the research context, confirmed the LLM search results, and verified all analysis and results. LLMs were also used for editing and typechecking this manuscript.

\end{document}